\documentclass[lettersize,journal]{IEEEtran}
\usepackage{amsmath,amsfonts}
\usepackage[normalem]{ulem}
\usepackage{color,soul}

\usepackage{graphics} % for pdf, bitmapped graphics files
\usepackage{epsfig} % for postscript graphics files
\usepackage{times} % assumes new font selection scheme installed
\usepackage{amsmath} % assumes amsmath package installed

\usepackage{amssymb}
\usepackage{yfonts}

\usepackage{algpseudocode}
\usepackage[ruled,vlined]{algorithm2e}

\usepackage{amsmath}

\usepackage{array}
\usepackage[caption=false,font=normalsize,labelfont=sf,textfont=sf]{subfig}
\usepackage{textcomp}
\usepackage{stfloats}
\usepackage{url}
\usepackage{verbatim}
\usepackage{graphicx}
\usepackage{cite}

\usepackage{booktabs}
\usepackage{footnote}
\usepackage{tabularx}
\usepackage{multirow}%提供跨列命令\multicolumn{}{}{}
\usepackage{hyperref}
\usepackage[flushleft]{threeparttable}

\begin{document}

\title{
% \LARGE \bf
Keypoint-Based Bimanual Shaping of Deformable Linear Objects under Environmental Constraints using  Hierarchical Action Planning%
}

\author{Shengzeng Huo, Anqing Duan, Chengxi Li, Peng Zhou, Wanyu Ma and David Navarro-Alarcon%
\thanks{This work is supported in part by the Key-Area Research and Development Program of Guangdong Province 2020 under project 76, in part by the Research Grants Council of Hong Kong under grants 14203917 and 15212721, in part by the Jiangsu Industrial Technology Research Institute Collaborative Research Program Scheme under grant ZG9V, and in part by The Hong Kong Polytechnic University under grant 8B01}% <-this % stops a space
\thanks{All authors are with The Hong Kong Polytechnic University, Department of Mechanical Engineering, Hung Hom, Kowloon, Hong Kong.}}



\maketitle
% \thispagestyle{empty}
% \pagestyle{empty}

%page
% \thispagestyle{fancy} % IEEE模板在\maketitle后会自动声明\thispagestyle{plain}，
%                             % 导致第一页什么都没有。所以得把plain更改为fancy
% \lhead{} % 页眉左，需要东西的话就在{}内添加
% \chead{} % 页眉中
% \rhead{} % 页眉右
% \lfoot{} % 页眉左
% \cfoot{} % 页眉中
% \rfoot{\thepage} %页眉右，\thepage 表示当前页码
% \renewcommand{\headrulewidth}{0pt} %改为0pt即可去掉页眉下面的横线
% \renewcommand{\footrulewidth}{0pt} %改为0pt即可去掉页脚上面的横线
% \pagestyle{fancy}
% \rfoot{\thepage}

%%%%%%%%%%%%%%%%%%%%%%%%%%%%%%%%%%%%%%%%%%%%%%%%%%%%%%%%%%%%%%%%%%%%%%%%%%%%%%%%
\begin{abstract}
This paper addresses the problem of contact-based manipulation of deformable linear objects (DLOs) towards desired shapes with a dual-arm robotic system. To alleviate the burden of high-dimensional continuous state-action spaces, we model the DLO as a kinematic multibody system via our proposed keypoint detection network. 
% \textcolor{blue}{model...via keypoints or model...via network?}
This new perception network is trained on a synthetic labeled image dataset and transferred to real manipulation scenarios without conducting any manual annotations. 
% , which in turn requires the network of our perception system to be trained for keypoints detection only.
% Notably, the training image dataset used is synthetic and manual annotation-free.
Our goal-conditioned policy can efficiently learn to rearrange the configuration of the DLO based on the detected keypoints.
The proposed hierarchical action framework tackles the manipulation problem in a coarse-to-fine manner (with high-level task planning and low-level motion control) by leveraging on two action primitives. The identification of deformation properties is avoided since the algorithm replans its motion after each bimanual execution.
The conducted experimental results reveal that our method achieves high performance in state representation of the DLO, and is robust to uncertain environmental constraints. 
% To the best of the authors’ knowledge, we are the first one to utilize contact to deform DLOs through bimanual manipulation without pregrasping.
% \textcolor{blue}{Remark. 1) this abstract is re-ordered and I think some sentences could be deleted for concision; 2) Does \textcolor{red}{\href{https://www.youtube.com/watch?v=AJllD3AiSqs}{this work}} use pregrasping? Maybe cite it somewhere in Introduction.}
\end{abstract}


%%%%%%%%%%%%%%%%%%%%%%%%%%%%%%%%%%%%%%%%%%%%%%%%%%%%%%%%%%%%%%%%%%%%%%%%%%%%%%%%
\section{INTRODUCTION}
% Deformable objects are very common in our daily life, such as the charging cables of your phones. For decades, robotic manipulation has studied intensively on rigid objects. However, 
% Deformable object manipulation is a fundamental ability that robots should have for many real applications \cite{sanchez2018robotic}, range from flexible cables arrangement \cite{zhu2018dual}, folding cloth \cite{garcia2020benchmarking} to surgical robots \cite{navarro2016automatic}. 
\IEEEPARstart{D}{eformable} object manipulation has many promising applications in growing fields, such as flexible cable arrangement \cite{zhu2018dual}, clothes folding \cite{garcia2020benchmarking}, and surgical robots \cite{navarro2016automatic}. 
% Compared with planar or solid objects, deformable linear objects (DLOs), such as ropes, cables, and rods, attract more attention due to their simplicity \cite{sanchez2018robotic}.
Among them, manipulation of deformable linear objects (DLOs) attracts much attraction due to its relevance in several manufacturing industries \cite{galassi2021robotic}, such as wiring harness and knot tying \cite{sanchez2018robotic}.

Although great progress has been recently achieved in deformable object manipulation (e.g. \cite{navarro2013model, zhu2021vision, navarro2017fourier}), shaping DLOs with environmental contacts remains an open problem. Compared with rigid objects, this problem is much more challenging due to the complex physical dynamics of infinite degree-of-freedom DLOs. 
% Our insight is that, rather than a precise dynamic model of a DLO, the keypoints at each time step are more practical and crucial to plan the robot actions. The motivations of the insight are (1) the accurate dynamics model is difficult to establish; (2) human is capable of manipulating DLOs without system identification.
% \textcolor{blue}{The first motivation is repetitive and the second motivation is so universal. Here goes my suggested version for this part:}
Our strategy is that instead of analytic physical dynamics, the DLO modeling is simplified to a kinematic multibody featured by several keypoints. The assumptions of our strategy are 1) the keypoint representation is sufficient for the contact-based shape matching problem, and 2) the shape error incurred by the modeling simplicity can be compensated for by coarse-to-fine manipulation.
% do not have the accurate physical model about DLOs but we still are able to finish the tasks. In this paper, 
% Instead of establishing precise dynamic models of DLOs, we pay attention on their keypoints to guide our hierarchical action planning. 
This paper aims to develop a complete algorithm (including perception and planning) to tackle the task of contact-based shaping of DLOs with bimanual manipulation. 
%Based on our keypoint detection in perception, our system 
% In order to keep the stable state of DLO, humans always use contacts to limit the DoF of it. However, it is very difficult to give robots similar capabilities, especially under the condition of the lack of tactile and force sensing. Firstly, although this task can be formulated as a classical pick-and-place problem, the selected picking points is infinite and the motion path should be designed carefully. Secondly, this task is a long-horizon task with multi-steps. Each action sequences should try to minimize the local error and consider the global state to avoid dragging in local minimum at the same time. The goal of this paper is to develop a feasible solution to this problem.

%\subsection{Related Work}
% a) \textit{Manual Hard-coded Feature Descriptor:}
%a) \textit{Perception:}
Many researchers have worked on the representation of DLOs in vision \cite{zhu2021challenges}.
% Vision is the most straightforward representation about the global deformation of objects \cite{zhu2021challenges}. 
Angles \cite{wang2018unified} and curvatures \cite{navarro2014visual} are intuitive hard-coded descriptors for shape feedback, whose generalization is poor. \cite{zhu2018dual,navarro2017fourier} develop Fourier-based descriptor; however, they require high computation cost during online perception.
% these methods require high dimensional vectors to achieve high accuracy. Although \cite{hu20193} and \cite{zhu2020vision} use Principle Component Analysis (PCA) \cite{wold1987principal} to reduce the dimension, the vectors in the sub-space are not semantic and explanable. 
% b) \textit{Data-driven Methods:} 
% Data-driven based shape analyses \cite{xu2016data}, \cite{zhang2008deformation} have gain in popularity as it offers a useful alternative to model-based approaches. Latent space approaches have recently achieved many successful results in image analysis \cite{hoff2002latent}, due to its capability to encode high-dimensional data into a meaningful internal representation. 
Data-driven based shape analysis has gained popularity in feature extraction \cite{zhou2021lasesom}.
% has successfully using latent code for soft object semantic analysis. However, these methods hold the hypothesis that the deformable objects can be completely observed, which is difficult to achieve in DOM and this work is not general enough.
\cite{tang2018framework} employs the Gaussian Mixture Model for its physics simulation engine, assuming the physical model of the deformable objects is known. \cite{tanaka2018emd} proposes an Encode-Manipulate-Decode network for cloth manipulation. However, it needs tremendous data collection and the latent vector is not semantic.
% \cite{yan2020self} uses iterative STNs to divide the ropes into multiple segments with self-supervised training. However, this state estimation procedure still needs numerous data collection to finish the transfer from rendered images to real data.
% b) Data-driven Feature
% b) Occlusion is a very common phenomenon in vision-based robotic manipulation tasks. In DOM, fully-observable is very difficult to achieve since the state of the deformable object is complex and self-occlusion usually occurs. Hence, a occlusion-robust deformable object tracking framework is necessary in DOM. Although \cite{schulman2013tracking} tracks deformable objects bases on point cloud data, it requires a generic physics simulation. \cite{chi2019occlusion} presents an occlusion-robust RGBD sequence tracking framework. However, this method relies on time-series observation and the deformation is not complex and in 2D. 
% b) \textit{Synthetic Simulation-to-real:}
% Recently, machine learning has been widely adopted in the field of robotics. 
% Recently, machine learning drives the development of deformable object manipulation. 
Since real data is expensive to collect, learning on synthetic datasets and transferring to physical situations is an alternative solution \cite{johns2016deep}. 
% Among them, GraspIt! simulator \cite{miller2004graspit} serves as a public platform for grasp synthesis. For deformable objects, 
\cite{seita2020deep} simulates 2D fabric smoothing on a mesh grid connected by various springs. \cite{sundaresan2020learning} forms a braid of rope through twisting cylindrical meshes; this work needs a sphere mesh on one end to break out the symmetry.
\cite{yan2020self} generates images with a random b-spline curve with six control points; however, it still needs a real dataset for perception finetuning.
% from rendered images to real data.
% b) \textit{Deformable Linear Object Manipulation:}
% Model-based and model-free are two main branches about the control of deformable object manipulation. Mass-Spring \cite{essahbi2012soft} and Finite-Element (FEM) models \cite{kaufmann2009flexible} are two kinds of dynamic models. However, their performance heavily rely on the accurate prior knowledge of the deformables \cite{mcconachie2020manipulating}. 
% Compared with planar or solid objects, deformable linear objects (DLOs), such as ropes, cables, and rods, attract more attention due to their simplicity \cite{sanchez2018robotic}. 

Robotic manipulation of deformable objects has been studied with various formulations and assumptions, including model-based and model-free approaches. With the pregrasping hypothesis, \cite{zhu2018dual,jin2019robust} consider the deforming task as shape servoing and approximate the local deformation model with a linear Jacobian matrix, while the global convergence is not guaranteed. Formulating the task as a multi-step pick-and-place manipulation problem, \cite{sundaresan2020learning,yan2020self} conduct the tasks with single-arm policy while real data collection is required for sim-to-real transferring or human visual demonstration. \cite{tang2018track} assembles DLOs for specified fixtures with dual robots,
% \sout{but they do not take contacts into consideration} 
yet contacts are not taken into consideration. 
% Instead of estimating the complex global model, model-free approaches replace it as local linear Jacobian matrix to plan, such as \cite{navarro2013model} and \cite{navarro2016automatic}. However, the global convergence is hard to guarantee due to the lack of global view.
% c) \textit{Task and Motion Planning}:
Task and motion planning (TAMP) is a solution to tackle this multi-step decision-making task \cite{yin2021modeling} through factorizing \cite{5980391} the planning process into discrete symbolic reasoning and continuous motion generation \cite{wells2019learning}.
However, a majority of TAMP algorithms assume rigid objects, whose predictable dynamics are not available in deformable objects, let alone under contact constraints. 
% Note that we choose to ignore the complex physical dynamics and focus on its model. Rather than planning the whole long-horizon sub-goals, we re-plan our next sub-goal after each action.   
% \cite{dantam2018incremental} leverage incremental constraint solving to efficfiently incorporate geometric information at the task level. Our work based on this theory to develop incremental re-planning paradigm in task level to avoid the acquisition of the complex non-linear dynamics.

%\subsection{Our Contribution}
\cite{zhu2019robotic} exploits environmental contacts for manipulation of DLOs, which is achieved with some customized mechanical grippers and the assumption of pregrasping. We advance the achievement to manipulating DLOs from arbitrary configurations to the desired goal states under environmental constraints. The shape of the DLO is characterized with a sequence of ordered keypoints, an approach that narrows the state-action search space for bimanual manipulation. To deal with the complex contact configurations, a coarse-to-fine planning framework with two defined action primitives is derived. 
% Instead of modeling the complex physical dynamics of DLOs, we focus on the state estimation and the sub-goal planning in next time-step, which is similar to the manipulation behaviors of humans. 
The original contributions of this work are as follows:
% \begin{itemize}
%     \item 
%     Learn to detect the keypoints from visual representations of DLOs based on the synthetic dataset without requiring any manual annotation.
%     \item Propose a TAMP hierarchical control framework for contact-based DLOs bimanual manipulation in a coarse-to-fine manner.
%     \item Demonstrate various contact-based DLOs bimanual manipulation experiments in real environments.
% \end{itemize}
\begin{itemize}
    \item 
    A novel data-driven perception approach for DLOs whose network is trained on a synthetic dataset.
    % without requiring any manual annotation.
    % to render a synthetic visual dataset for supervised feature extraction of DLOs without requiring any manual annotation.
    \item A hierarchical action planning framework for shaping DLOs under environmental constraints in a coarse-to-fine manner.
    \item Experimental results to validate our solution for contact-based DLOs bimanual manipulation in real environments.
\end{itemize}

%\subsection{Paper Organization and Notation}
The remainder of this paper is organized as follows. Sec. II states the task's formulation. Sec. III explains the perception. Sec. IV reports the planning framework. Sec. V reports the results and Sec. VI gives the conclusions. 

\section{PROBLEM FORMULATION}
\begin{figure}
    \centerline{\includegraphics[width=\columnwidth]{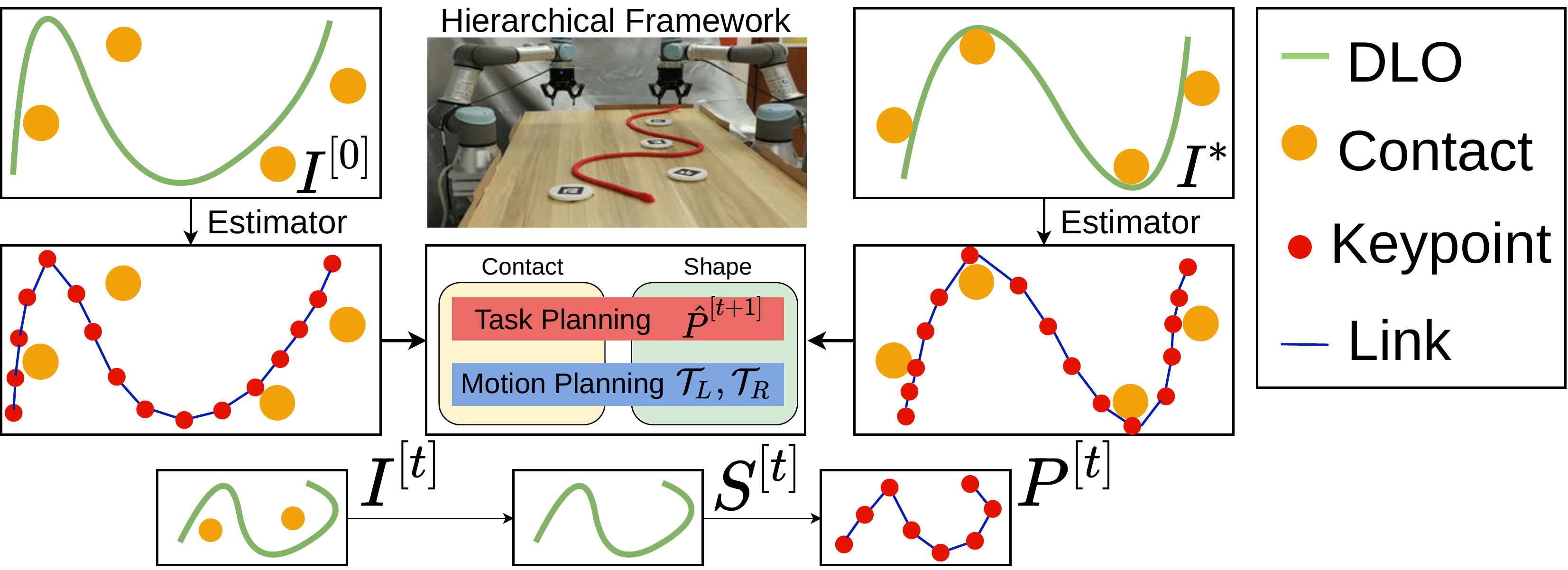}}
    \caption{Overview of the keypoint-based bimanual manipulation framework for shaping DLOs with contacts. Given the goal $I^*$, the DLO manipulation is formulated as a goal-directed task from the initial configuration $I^{[0]}$. At each time step $t$, the perception detects the sequential keypoints $P^{[t]}$ corresponding to the state of the DLO $S^{[t]}$ extracted from the visual observation $I^{[t]}$. The hierarchical control framework takes the current $P^{[t]}$ and the goal $P^{*}$ keypoints as input and outputs the action plan ($\mathcal{T}_L,\mathcal{T}_R$). The whole algorithm replans based on the new observation after the execution of the robots. The scheme iterates until reaching the desired goal.
    % \textcolor{blue}{we may also forget about $I^{[0]}$ in the figure so that $I^{[t]}$ is used consistently.}
    % \textcolor{magenta}{Suggestions 1)for colored curves and dots, explain what each color means; 2)make sure all the quantities appear before introducing the fig in the text (also all other figs)}.
    }
    \label{1-overview}
\end{figure}

The architecture of our vision-based manipulation system is depicted in Fig. \ref{1-overview}. Given a goal observation $I^{*}$, our task is to manipulate the DLO with an initial configuration $I^{[0]}$ to match it. Assuming the DLO has an obvious color contrast with the background, we segment the state of the DLO $S^{[t]}$ from a raw image $I^{[t]}$ with a color filter. To simplify the problem, we consider circular contacts with known size $C=\{c_1,\cdots,c_k,\cdots,c_q\}$ in the observation $I^{[t]}$.

% Given the initial vision environment $I^{[0]}$ and the state $S^{[0]}$, our goal is to deform the deformable linear object to a desired configuration $I^{*}$, which is proposed by a vision observation $I^{*}$. Beside the deformable object, the vision observation $I^{[t]}$ also involves the contacts $C$, which is the helper to shape the ropes.
Formulating deformable object manipulation as a multi-step decision-making process, our aim is to obtain an action plan $\mathcal{A}=(A^{[1]},\cdots,A^{[t]},\cdots,A^{[H]})$ within $H$ steps, such that the last state $S^{[H+1]}$ (with the transition function $S^{[H+1]}=A^{[H]}\times S^{[H]}$) reaches the goal state $S^*$. To apply TAMP framework for this challenging task, we make some modifications versus perception and planning.
% , where each $a^{[t]}$ in the task language $\mathfrak{L}$ to achieve the goal state $s^*$. 
% Specifically, $A=(a^{[1]},\cdots,a^{[t]},\cdots,a^{[h]})$, where each $a^{[t]}$ represents the action of the robotic manipulators at time step $t$. Hence, a concrete transition at step $t$ from $s^{[t]}$ to $s^{[t+1]}$ is represented as $s^{[t+1]}=a^{[t]}(s^{[t]})$. After the whole horizon $h$, the last transition $s^{[H+1]}=a^{[H]}(s^{[H]})$.
% Due to the infinite degree of freedom of DLO, we propose a hierarchical TAMP framework for robotic bimanual manipulation. Since most of TAMP algorithms hold the hypothesis of rigid objects, we make some modifications to fit our task. 
The state $S^{[t]}$ of the DLO is depicted as $S^{[t]}=\{s_1^{[t]},\cdots,s_i^{[t]},\cdots,s_n^{[t]}\}$. Based on the kinematic multibody model \cite{wnuk2020kinematic}, we describe the DLO $S^{[t]}$ as a list of sequential keypoints $P^{[t]}=\{p_1^{[t]},\cdots,p_j^{[t]},\cdots,p_m^{[t]}\}$ ($m\ll n$) since it allows us to 1) narrow down the search space from high-dimensional state to low-dimensional latent space, and 2) obtain a compact feedback vector for semantic bimanual manipulation. Note that the end of the DLO closer to the left robot is denoted as the first keypoint in the perception.
% 1) reduce the dimension of the shape representation and 2) extract the fundamental descriptors for the following policy. 
Based on the description, our hierarchical control framework combines high-level task planning and low-level motion control. Taking $P^{[t]}$ and $P^*$ as input, the high-level model designs the sub-goals $\hat{P}^{[t+1]}$, while low-level model plans the local motion $A^{[t]}$ to achieve $\hat{P}^{[t+1]}$. Note that $\hat{P}^{[t+1]}$ is the designed sub-goal and is different from the detected keypoints $P^{[t+1]}$ at time step $t+1$.
% Two action primitives are defined to implement coarse-to-fine shape matching.
We choose bimanual manipulation in a tabletop environment instead of a single-arm to 1) constraint the unpredictable displacement of the DLO, and 2) enrich the diversity of the action. In this case, each plan $A^{[t]}$ is defined as $A^{[t]}=[\mathcal{T}^{[t]}_L\ \mathcal{T}^{[t]}_R]$, where $\mathcal{T}^{[t]}_L$ and $\mathcal{T}^{[t]}_R$ are the actions of the left and right arm, respectively, including motion, grasping, and releasing.

\section{Perception}
Our perception takes the visual binary image $S^{[t]}$ as input and outputs the corresponding keypoints $P^{[t]}$. To avoid time-consuming real-world data collection for training, we render an annotated synthetic image dataset for supervised learning (Sec. III-A) and finetune the output of the network through the geometric constraints (Sec. III-B).

\subsection{Synthetic Dataset Generation}
\begin{figure}
    \centerline{\includegraphics[width=\columnwidth]{"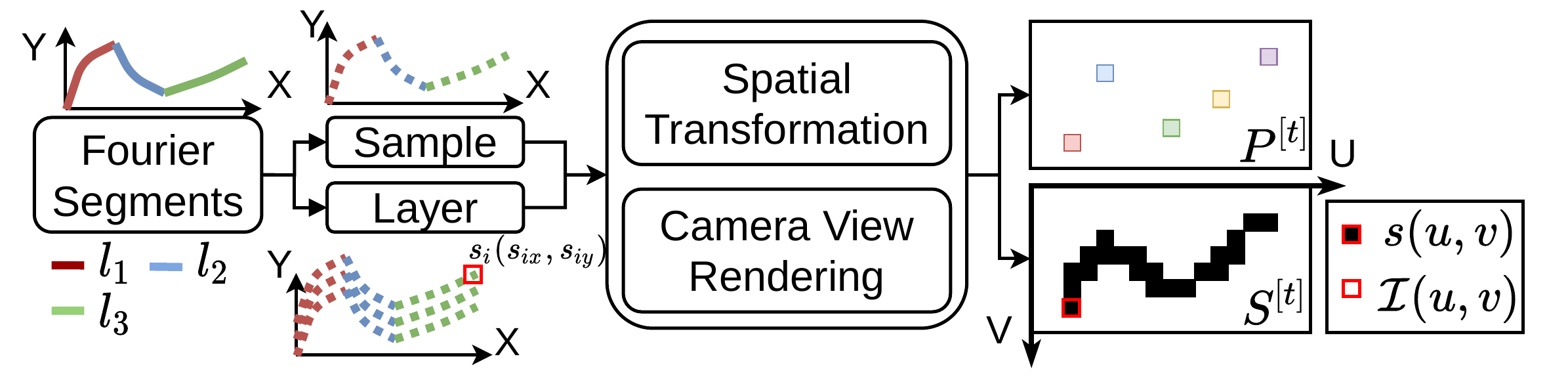"}}
    \caption{Data generation of labeled synthetic DLOs. Base on the Fourier series, we generate multiple curve segments and concatenate them end-to-end. This raw data undergoes sampling and stacking for keypoint labels and image input, respectively. After the spatial transformation for data augmentation, we simulate image rendering based on the camera acquisition principle.
    % \textcolor{magenta}{Suggestion. Please define the X and Y axis in a coordinate.}
    }
    \label{2.1-data_generation}
\end{figure}
% In order to extract precise structured shape descriptors of DLOs, we desire sequential keypoints $P^{[t]}$ since 1) narrow down the search space from high-dimensional state to low-dimensional latent and 2) obtain a feedback latent representation for semantic bimanual manipulation. We leverage a learning-based feature extractor due to its robustness and efficiency. 
In this section, we simulate DLOs to facilitate the keypoint detection $P^{[t]}$ from the binary image $S^{[t]}$, as illustrated in Fig. \ref{2.1-data_generation}. 
% propose to synthesize a simulated visual dataset about the DLO. To minimize the discrepancy between the synthesis and the real data ($S^{[t]}$, $P^{[t]}$), we generate the dataset in the format of 2D images automatically, as illustrated in Fig. \ref{2.1-data_generation}.\textcolor{blue}{Sidenote it is not clear how 2D format images contribute to the data discrepancy minimization.}
% We avoid the data collection through rendering a image dataset. To minimize the additional effort for transferring from the synthetic dataset to real situations, their similarity should be as high as possible.
% Assuming the high color contrast with the tabletop environment, we extract the region of DLO $S$ from the raw input $I$ via color-based segmentation, representing as a binary image. Hence, our state estimation problem is simplified as keypoints detection within DLOs. To bridge this gap, we construct a labeled dataset for learning.
Geometrically, a DLO refers to an object whose length is much larger than the diameter of its cross-section. Thus, we mathematically describe it as a continuous curve.
% To generate synthetic training data, we adopt a numerical model to describe their shape.
% Taking the deformation of the ropes into consideration, we use several segments to model the shape of them instead of a whole body. 
Taking the deformation of DLOs into consideration, our model utilizes several 2D curve segments based on the Fourier series \cite{navarro2017fourier} to depict the local shape of DLOs $(l_1,l_2,\cdots)$, where each curve segment $l$ is described along the X-axis:
% The interested area $I$ is represented as a binary image, its point cloud $S=\{s_1,\cdots s_i\}$ is acquired according to the internal parameters of the depth camera.
% However, with this observation, we still can't obtain the state of the object. Data-driven feature descriptors are more generilizable than manual hard-coded features, which enquired high cost labeled data. This problem is more prominent in our task since multiple numerical accurate labels instead of single perceptual knowledge are required. In order to deal with this issue, we propose a synthetic data generation method to prepare the labeled dataset.To achieve a dataset, we need a numerical method to generate deformable linear objects. The most straightforward method to describe the shape is continuous spatial curve along the skeleton line in Euclidean space $\boldsymbol{f}(s): \mathbb{R} \mapsto \mathbb{R}^{3}$ \cite{wnuk2020kinematic}. Mathematically, Polynomial Parameterization, Bezier curve, 
% NURBS curve \cite{piegl6nurbs} are some models to depict continuous curves. 

% \begin{equation}
% y_i=f(x_i)=a_i+b_i\sum_{i=1}^n\cos(c_i\cdot i)+d_i\sum_{i=1}^n\cos(e_i\cdot i)
% \end{equation}
\begin{equation}\label{fourierdef}
y=f(x)=\frac{a_{0}}{2}+\sum_{n=1}^{N}\left[a_{n} \cos (n \omega x)+b_{n} \sin (n \omega x)\right]
\end{equation}
% where $a$...
where $a_0$ is the bias of the Fourier descriptor at zero frequency and $N$ is the number of harmonics under consideration. The coefficients of the n-th harmonic $a_n,b_n$ are defined as $a_{n} =\frac{2}{T} \int_{t_{0}}^{t_{0}+T} f(t) \cos (n \omega t) d t $ and $b_{n} =\frac{2}{T} \int_{t_{0}}^{t_{0}+T} f(t) \sin (n \omega t) dt$, respectively, where $\omega$ denotes the frequency. Note that the discrete points of the curve $l$ are in order along the X-axis with this definition.
% \textcolor{magenta}{Remark. 1) Is the index $t$ different from the one (say time step) introduced previously? 2) Is $f(t)$ in \eqref{fourierdef} the same as the one used for defining $a_{n}$ and $b_{n}$?}
% This Fourier-based model provides the source for the following image rendering shown in Fig. \ref{2.1-data_generation}. 
Each DLO $L$ consists of several end-to-end connected Fourier series-based segments $L=(l_1,l_2,\cdots)$ and the point of it is represented as $s_i=(s_{ix},s_{iy})$. Next, we simulate the raw input $S^{[t]}$ and our desired keypoints $P^{[t]}$, respectively. We denote $m$ keypoints from $S^{[t]}$ in a coarse-to-fine manner. Initially, $m$ candidates are sampled uniformly according to Euclidean distance. Since the points with high curvature describe the contour of the DLO, we also desire those as keypoints. The curvature of a point $s_i$ is defined as:
\begin{equation}
    \alpha_i=\left<f_{-}'(s_i),f_{+}'(s_i)\right>
\end{equation}
% \begin{equation}\label{arccosfunc}
%     \alpha_i=\arccos(f_{-}'(s_i)\cdot f_{+}'(s_i)/(|f_{-}'(s_i)|\cdot |f_{+}'(s_i)|))
% \end{equation}
% \textcolor{magenta}{$\arccos$ only takes one argument.}
where $f_{-}'(s_i)=\Vec{s}_i-\Vec{s}_{i-1}$ and $f_{+}'(s_i)=\Vec{s}_{i+1}-\Vec{s}_{i}$. Here, $\left<\Vec{a},\Vec{b}\right>$ denotes the function about computing the angle between two vectors $(\Vec{a},\Vec{b})$. According to this definition, we substitute the points whose curvatures are larger than a threshold $\tau_u$ for their corresponding nearest uniform candidates, enabling the number of the keypoints keeping a constant $m$.
% \textcolor{blue}{replace...with what? and near to what and what does a candidate refer to?}
% Specifically, we include the points whose curvatures are larger than a pre-defined threshold $\tau_u$ and replace \textcolor{blue}{replace...with what?} the nearest candidates of the previous uniform sampling \textcolor{blue}{near to what and what does a candidate refer to?}, enabling the number of the keypoints keeping a constant $m$.
% the absorbed points replace the uniform points nearest to it and the number of keypoints remain the same for each sample.
% \textcolor{blue}{This part regarding $P^{[t]}$ is hard to follow. E.g. what does "absorb" mean? Can this mechanism be described mathematically?}
% the direction change between the following vector and the previous one. 
For $S^{[t]}$, we stack the curve $L$ along Y-axis to simulate the cross-section of the DLO. 
% Note that center-line extraction, or skeletonization is also a challenge in state estimation. 
After these steps, both the sampled keypoints and the stacked layers enter into spatial transformation for data augmentation and camera view rendering for image processing. Spatial transformation, including translation and rotation, is significant for balancing the distribution of samples. Camera view rendering consists of resizing the curve into the region of interest and reorder of the points into an image format. Since we adopt a binary image $S^{[t]}$ to represent the DLO, the pixel at $S(u,v)$ is positive if any point locates within its surroundings:
% Our generation pipeline also involve arbitrary planar transform for data augmentation. The final step is camera view rendering. Firstly, we ought to resize and translate the original curve into the region of interest of the camera. Next, since the raw input and the label are both binary images, we iteratively line-to-line to generate the images while the pixel is positive if points exist. 
\begin{equation}\label{pixeldef}
    S(u,v)=\left\{
    \begin{aligned}
        1,\quad & \exists s_i\in \mathcal{I}(u,v) \\
        0,\quad & otherwise
    \end{aligned}
    \right.
\end{equation}
where $s_i\in \mathcal{I}(u,v)\Longleftrightarrow\{ w_u<s_{ix}<w_{u+1}\}\cap\{ h_v<s_{iy}<h_{v+1}\}$, $u$ and $v$ are the horizontal and vertical position of the pixel in the image, respectively. 
% \textcolor{blue}{Consider to define $w_v$ etc. The equation with $\Longleftrightarrow$ is not clear to me. Do you intend to define $\mathcal{I}$?}
For the labeled keypoints, we transform them from Cartesian frame to image frame, represented as $p_j(u_j,v_j)$ in sequence. 

With this generation pipeline, we obtain binary images describing the shape of DLOs and their corresponding annotated sequential keypoints. 

% we render images similar to the real observation of common cameras and extract the pixels of the corresponding keypoints in the image frame. This dataset offers a basis for feature estimation.
% Finally, we acquire two binary images as input and corresponding label respectively.

\subsection{Keypoint Detection}
% Compared with the raw high-dimensional sensory observations, keypoints are more effective to represent the object for planning and control. 
\begin{figure}
    \centerline{\includegraphics[width=\columnwidth]{"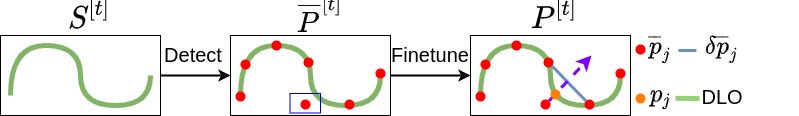"}}
    \caption{Illustration of the geometric finetuning. The point locating on the background area is revised along the direction vertical to its tangent.
    % \textcolor{magenta}{Remark. 1) what is a point's tangent? 2) by $\nabla$ do you mean $\delta$?}
    }
    \label{2.2-fine_tune}
\end{figure}

To detect the keypoints from the visual image $S^{[t]}$, we design a network to predict the keypoints of DLOs. More details about the network structure and the training process are discussed in Sec V.

While the network is generalizable across different shapes of DLOs, errors are still unavoidable. As illustrated in Fig. \ref{2.2-fine_tune}, some outputs visually locate on the area of the background, which conflicts with the prior knowledge that the keypoints locate within the DLO. Hence, we consider this geometric constraint finetuning.
% In our case, the binary images and coordinates are the input and the output of the network respectively. The output coordinate probably unreasonably locates on the background area, shown in Fig. \ref{2-Fine tune}.
% In order to deal with this issue, we propose a geometry-based keypoints finetuning scheme. 
For an output $\overline{p}_j=(\overline{u}_j,\overline{v}_j)$ that fails, namely $S^{[t]}(\overline{u}_j,\overline{v}_j)=0$, we utilize the adjacent pixels $(\overline{p}_{j-1},\overline{p}_{j+1})$ to correct it, which is divided into two cases: (1) the ends are adjusted to the nearest pixels in the area of DLO and (2) the intermediate keypoints are revised through searching along the direction vertical to its tangent space $\delta \overline{p}_j$:
% We introduce geometric priors to deal with this issue. For each output coordinate, we judge if it locates on the interested area of the deformable object. If it is not, we refine the position within its neighbourhood. For two corners, we search the nearest pixels in the interested area around it. 
% For midpoints, we find the cross point along the direction vertical to the direction vectors.
% \begin{equation}
% \begin{aligned}
%     \Vec{\mathbf{f}}\cdot\overrightarrow{p_{i-1}p_{i+1}}&=0\\
%     f_x\cdot(y-p_{iy})&=f_y\cdot(x-p_{ix})
% \end{aligned}
% \end{equation}
% \begin{equation}
% \begin{aligned}
% &find\quad s_i(u_i,v_i)\\
% &\begin{array}
% {r@{\quad}r@{}l@{\quad}l}
% s.t.
% &s_i(u_i,v_i)=1\\
% &\overrightarrow{s_i \overline{p}_i}\cdot\delta \overline{p}_i=0 \textcolor{blue}{\text{ bracket not needed}}\\
% \end{array} 
% \end{aligned}
% \end{equation}
\begin{equation}
\begin{aligned}
&find\quad s_i(u_i,v_i)\\
& \begin{array}
{r@{\quad}r@{}l@{\quad}l}
s.t.
&S^{[t]}(u_i,v_i)=1\\
&\overrightarrow{s_i \overline{p}_j}\cdot\delta \overline{p}_j=0\\
\end{array} 
\end{aligned}
\end{equation}
where its tangent space $\delta \overline{p}_j$ is defined as $\delta \overline{p}_j=\Vec{\overline{p}}_{j+1}-\Vec{\overline{p}}_{j-1}$. 
% \textcolor{blue}{For the constraints, why it's not $S(u_i,v_i)=1$ and $\overrightarrow{s_i \overline{p}_i}\cdot\delta \overline{p}_i=0$?}
Notably, we denote $P^{[t]}$ as the finetuning result of the raw output $\overline{P}^{[t]}$.
% After the finetuning, the corresponding keypoint $\hat{p}_i$ is refined as $p_i$. Through the geometric refinement, our state estimation model is more robust to deal with noise in real data perception.

\section{Hierarchical Action Planning}
% \begin{figure}
%     \centerline{\includegraphics[width=\columnwidth]{"figures/3.1-tamp_structure.jpg"}}
%     \caption{Conceptual representation of the hierarchical bimanual manipulation setup. (a) Experimental setup. (b) Taking the predicted keypoints of the current state and goal state, we select the grasp points of dual-arms respectively and plan their corresponding motion.\textcolor{blue}{It seems the color for $P^*$ and $P^{[t]}$ should be exchanged as the goal state should be the one learning the yellow contact points. In addition, it is expected for this fig to convey more specific info than fig 1. Is it possible to explain the mechanism of $\hat{P}^{[t+1]}$ as this seems the emphasis of the text?} }
%     \label{3.1-tamp_structure}
% \end{figure}
% \textcolor{blue}{I really think this section should appear before Sec. Synthetic Keypoint Detection. The hierarchical architecture gives the blueprint of the paper and echoes the tile while technical details on dataset generation is comparatively trial and shall be considered to introduce later.}

\begin{figure}
    \centerline{\includegraphics[width=\columnwidth]{"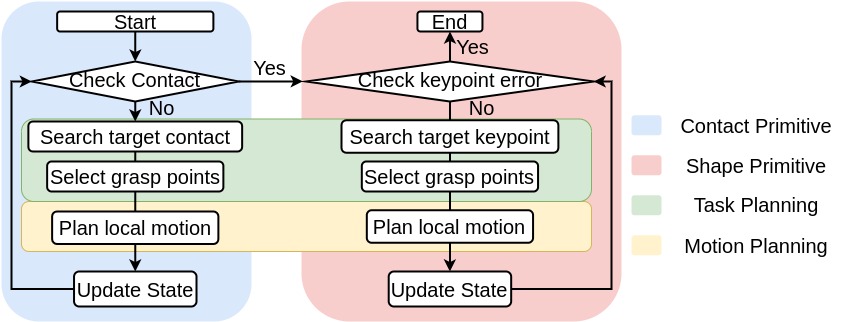"}}
    \caption{Flow chart depicting the conversion and details of the action primitives. }
    \label{3.2-tamp_algorithm}
\end{figure}

% We propose a hierarchical action planning framework for the task, outlined in Fig. \ref{3.1-tamp_structure}. Interleaving high-level task planning and low-level motion control, we define the task $\hat{P}^{[t+1]}$ concerning the detected keypoints ($P^{[t]}$,$P^*$) and derive a search-based action plan ($\mathcal{T}_L$,$\mathcal{T}_R$) 
% at each time step $t$.\textcolor{red}{
We propose a hierarchical action planning framework interleaving high-level task planning, where a sub-goal $\hat{P}^{[t+1]}$ is designed given the detected keypoints ($P^{[t]}$,$P^*$), and low-level motion control, where a search-based action plan ($\mathcal{T}_L$,$\mathcal{T}_R$) is derived 
at each time step $t$. Specifically, the design task $\hat{P}^{[t+1]}$ investigates the interested keypoints specification (selection and placement) of dual arms respectively, whereas ($\mathcal{T}_L$,$\mathcal{T}_R$) is an action plan to manipulate the DLO to the detailed configuration.
% The schematic illustration of our proposed framework is shown in Fig. 4.
% \textcolor{blue}{why brackets?}.
% For each action $A^{[t]}=[a^{[t]}_L\ a^{[t]}_R]$, we design the high-level sub-goal $S^{[t+1]}$ (the keypoints of pick and place) and plan the low-level motion $(\mathcal{T}_L,\mathcal{T}_R)$ to achieve the goal (the control of moving, grasping and releasing, considering the action sequence, obstacles, robotic kinematic and reachability). 
% Without any dynamics modeling, our framework re-plan after each step instead of forecasting.
% The high-level model designs the following sub-goal $P^{[t+1]}$ and selects the grasping points of dual-arms respectively based on the current state $P^{[t]}$ and the target estimated state $P^*$. The low-level model grasps the chosen points and plan the local motion to reach the desired sub-goal $P^{[t+1]}$. 
Under this framework, we define two multi-step action primitives, the contact primitive and the shape primitive to implement the task in a coarse-to-fine manner. The switch between them depends on the analysis of the contact constraints, as illustrated in Fig. \ref{3.2-tamp_algorithm}. 
% The whole algorithm iterates until reaching the desired goal shape with a small error. 
Sharing the same classical pick-and-place manipulation configuration between the primitives, we first detail the contact primitive (Sec. IV-A) and highlight the difference of the shape primitive (Sec. IV-B) afterward.

% Task and motion planning (TAMP) is an efficient solution to deal with long-horizon robotic manipulation tasks. Illustrated in Fig. \ref{3-TAMP structure}, task planning refers to a sub-goal planning and motion planning refers to the moving path of dual arms to achieve the sub-goal. Without an accurate dynamic model, we re-plan after each motion. Focusing on contact-based deformation issue, we develop a coarse-to-fine manner involving two modes, contact mode and shape mode.

% \begin{algorithm}
% 	\caption{Incremental TAMP}
% 	\label{TAMP}
% 	\textbf{Input} \hspace*{\algorithmicindent} $E^{[1]},E^*$\\
% 	\textbf{Output}\hspace*{\algorithmicindent} $A=\{a^{[0]},\cdots a^{[H]}\}$, $S^{[H+1]}$\\
% 	\textbf{Routine}
% 	\begin{algorithmic}[1]
% 	\STATE $S^{*},I^{*},P^{*},C,\mathcal{C},\mathcal{P}\leftarrow$ Configure the environment ($E^{*}$)
% 	\STATE $S^{[t]},I^{[t]},P^{[t]}$ Sense Environments ($E^{[t]}$)
% 	\WHILE{Contact Check ($S^{[t]},C,\mathcal{C}$)}
% 	    \STATE $c^s,c^e \leftarrow$Check Contact range ($S^{[t]},C,\mathcal{C}$)
% 	    \STATE $T^L,T^R\leftarrow$ Plan contact grasp ($C^s,C^e,P^{[t]},P^*$)
% 	    \STATE $a_L^{[t]},a_R^{[t]},\psi\leftarrow$ Plan motion ($T^L,T^R,P^{[t]},P^*,\mathcal{P}$)
% 	    \STATE $S^{[t+1]},P^{[t+1]}\leftarrow$ System Dynamics($S^{[t]},a_L^{[t]},a_R^{[t]},\psi$)
% 	   % \STATE $P^{[t+1]}\leftarrow$Update keypoint ($S^{[t+1]}$)
% 	\ENDWHILE
% 	\REPEAT
% 	    \STATE 	$T^L,T^R\leftarrow$ Search Max error ($P^{[t]},P^*$)
% 	    \Shttps://www.overleaf.com/projectTATE$a_L^{[t]},a_R^{[t]}\leftarrow$ Plan shape motion($P^{[t]},P^*,C$)
% 	    \STATE $S^{[t+1]},P^{[t+1]}\leftarrow$ System Dynamics($S^{[t]},a_L^{[t]},a_R^{[t]},\psi$)
% 	   % \STATE $P^{[t+1]}\leftarrow$Update keypoint ($S^{[t+1]}$)
%     \UNTIL {$||\\bigtriangledown s||<\eta$}
% 	\end{algorithmic} 
% \end{algorithm}

\subsection{Contact Primitive}
\begin{figure}
    \centerline{\includegraphics[width=\columnwidth]{"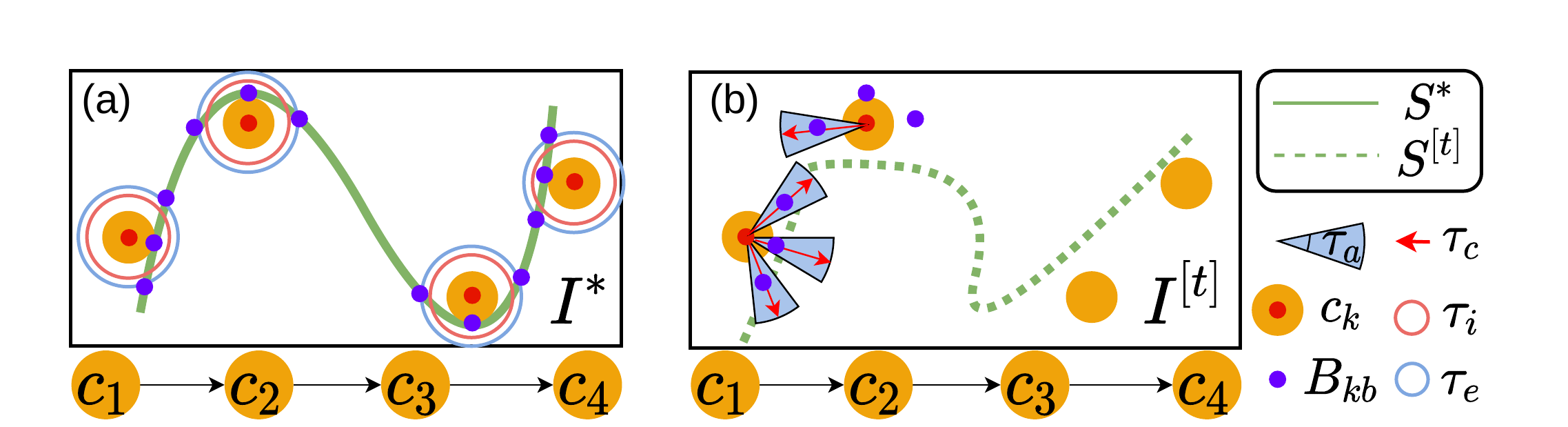"}}
    \caption{Graphical explanation of the contact primitive. (a) Based on the distance threshold $\tau_i$ and $\tau_e$, we acquire the benchmark $B_k$ in the goal state $S^*$ with respect to the contact $c_k$. 
    % \textcolor{blue}{if this is about benchmarks in the goal state, why it is not dashed green line?}
    (b) Based on the threshold $\tau_c$ and $\tau_a$, we search points that meet the conditions with respect to each benchmark $B_{kb}$.}
    \label{3.3-contact_primitive}
\end{figure}
\begin{algorithm}
	\caption{ContactPrimitive($P^{[t]},P^*,C,\mathcal{J},\mathcal{B},\mathcal{B}'$)}
	\label{ContactPrimitive}
	\While{$c_k\neq\varnothing$}
	{
	    $\mathcal{T}_L,\mathcal{T}_R\leftarrow$ GraspPlan ($P^{[t]},C,c_k,J$)\\
	   % $\mathcal{T}_R\leftarrow$ GraspPlan ($p^{[t]},C,c_k,J_{k}$)\\
	    \If{Reachable($c_k,B'_{k}$)}
	    {
	        $\mathcal{T}_R\leftarrow\mathcal{T}_R\cup$ Motion($c_k,\mathcal{B}'_{k}$)\\
	    }
	    \Else
	    {
	        $\mathcal{T}_L\leftarrow\mathcal{T}_L\cup$ Motion($c_k,\mathcal{B}'_{k}$)\\
	    }
	    $S^{[t+1]}\leftarrow$ System Dynamics $(\mathcal{T}_L,\mathcal{T}_R,S^{[t]})$\\
	    $P^{[t+1]}\leftarrow$Update keypoints $(S^{[t+1]})$\\
	    $c_k\leftarrow$ ContactSearch($P^{[t]},C,\mathcal{B}$)\\
	}
\end{algorithm}
Unlike the common robotic manipulation tasks, i.e., grasping, pushing, shape servoing, shaping DLOs under environmental constraints is a discrete-continuous mixed manipulation task, since contacts also provide external force for DLOs besides robots. With the state $S^*$ extracted from desired goal observation $I^*$, DLOs reach and stay in this goal configuration only if the contacts are constructed correctly. Hence, it is significant to construct the contacts first as coarse matching, denoted as contact primitive. The whole algorithm of this primitive is shown in Alg. \ref{ContactPrimitive}. 
% Specifically, the contact primitive includes selecting the target contact $c_k$ in high-level and planning the motion $(\mathcal{T}_L,\mathcal{T}_R)$ of the dual-arm robot to achieve it in low-level. We detail them respectively in the following and the whole algorithm of this primitive is shown in Alg. \ref{ContactPrimitive}. 

% Hence, it is necessary to arrange the configuration of DLOs around contacts. The prior work \cite{zhu2019robotic} introduce a notion called Angular Contact Mobility Index to represent the relative relations between the object and the contact. 
% To select the target contact $c_k$ at time-step $t$, we define a judgement criteria about whether the contact have been established.

Firstly, we analyze the role of contacts in shaping the DLO as $S^*$, as illustrated in Fig. \ref{3.3-contact_primitive}(a). Each contact $c_k$ supports its adjacent elements of the DLO, constraining its mobility. Hence, our algorithm validates the contact construction according to these elements. Specifically, we denote a set of benchmarks $\mathcal{B}=\{B_{kb}|k=1,\cdots,q,b=1,2,3\}$ for contact evaluation, which three benchmarks $(B_{k1},B_{k2},B_{k3})$ are defined for a contact $c_k$. Among them, the benchmarks ($B_{k1},B_{k2}$) are obtained through constrained optimization:
\begin{equation}
\begin{aligned}
B_{k1}=\arg\max_{s_i}||s_i&-c_{k}||_2,B_{k2}=\arg\max_{s_i}||s_i-B_{k1}||_2\\
s.t. \quad &\tau_i<||s_i-c_k||_2<\tau_e
\end{aligned}
\end{equation}
where ($\tau_i$,$\tau_e$) is distance threshold of the search area.
% establish a ring search area with respect to it,
% \begin{equation}
%     \mathcal{R}_k=\{r_1,r_2,\cdots,r_i\}
% \label{interet area}
% \end{equation}
% where $r_i\in\mathcal{R}_k\Longleftrightarrow \tau_i<||s_i-c_k||_2<\tau_e$ and ($\tau_i$,$\tau_e$) is distance threshold of the search area. 
% \textcolor{blue}{This definition is not clear, whats the relationship between $s_i$ and $r$? I think this definition shall be merged with the optimization problem below.}
% Within this region $\mathcal{R}_k$, we consider two benchmarks about the coverage of the contact $c_k$:
% \begin{equation}
% B_{k1}=\arg\max_i||r_i-c_{k}||_2,\ 
% B_{k2}=\arg\max_i||r_i-B_{k1}||_2
% \end{equation}
% \textcolor{blue}{The solution to the left-hand-side optimization problem seems to be the largest distance rather than benchmark positions. I think this shall be re-phrased.}
% \textcolor{blue}{Again, the solution to the optimization problem is an index of a circle rather than a benchmark position. Here goes my suggestion:}
% \textcolor{red}{
% \begin{equation}
% \begin{split}
% B_{k1}=\arg\max_{s_i}||s_i-c_{k}||_2\\
% s.t. \quad \tau_i<||s_i-c_k||_2<\tau_e\\
% B_{k2}=\arg\max_{s_i}||s_i-B_{k1}||_2\\
% s.t. \quad \tau_i<||s_i-c_k||_2<\tau_e
% \end{split}
% \end{equation}
% }
% \textcolor{blue}{The motivation for the formulation is that the optimization problem is constrained. The constraints not only lie on the search area, but also concern the goal shape.}
In addition, we also search for the nearest element $s^*_i$ of the goal shape $S^*$ to emphasize the support force from the contact $c_k$ to the DLO,
\begin{equation}
    B_{k3}=\arg\min_{s_i}||s_i-c_{k}||_2
\end{equation}
% Based on keypoint descriptors, we define a key area search paradigm to check if the contact met the condition or not, shown in Fig. \ref{3-contact mode}. 
% Note that we simplify the task by considering circular contacts with known radius on the table. From the view of the provided goal image, we first search the keypoints around the contact from the original binary input.
% Given the desired goal state $S^*$, we denote a set of benchmark $\mathcal{B}_k=\{B_{kb}|b=1,2,3\}$ respect to each contact $c_k$. 
% Firstly, we search the nearest point to the contact within the desired state of DLO $S^*$,
% \begin{equation}
%     B_{k1}=\arg\min_i||s_i-c_{k}||_2
% \end{equation}
% Second, we establish a ring search area for each contact $c_k$,
% \begin{equation}
%     \mathcal{C}_k=\{s_1,s_2,\cdots,s_i\}
% \label{interet area}
% \end{equation}
% where $s_i\in\mathcal{C}_k\Longleftrightarrow r_i<||s_i-c_k||_2<r_e$. Within $\mathcal{C}_k$, we denote two search benchmark:
% \begin{equation}
% B_{k2}=\arg\max_i||s_i-c_{k}||_2,\ 
% B_{k3}=\arg\max_i||s_i-B_{k2}||_2
% \end{equation}
% \textcolor{blue}{the same problem as aforementioned}
After the search, we re-order the benchmarks $B_{kb}\in\mathcal{B}$ along the sequential keypoints $P^*$. 
These benchmarks act as baselines to assess the contact construction. A benchmark $B_{kb}$ is satisfied if we find an element $s_i$ that fulfills two thresholds $\tau_c$ and $\tau_a$ (graphical explanation in Fig. \ref{3.3-contact_primitive}(b)): 
% \textcolor{blue}{$\tau_\alpha$ or $\tau_a$?}
% An element $s_i$ of the DLO that satisfy the benchmark $B_{kb}$ is required to fulfill two conditions for distance $\tau_c$ and direction $\tau_\alpha$ respectively:
% Given the state $S^{[t]}$ at time-step $t$, we search if there are points $s^{[t]}_i$ locate on the region of interest of each benchmark that meet the below conditions at the same time:
% With the search benchmark, we define the corresponding keypoint index $\mathcal{P}$:
% \begin{equation}
%     \mathcal{P}_k=\arg\min_i||p^*_i-\mathcal{C}_k||_2
% \end{equation}
% With the computed $\mathcal{C},\mathcal{P}$ according to the desired goal $E^*$, we check the contact configuration in Alg. \ref{ContactCheck}. For each key index, the judgement involves two necessary conditions:
\begin{equation}
 \begin{cases}||s_i-c_k||_2&<\tau_c\\
 \left<\overrightarrow{c_k s_i},\overrightarrow{c_k B_{kb}}\right>&<\tau_a
 \end{cases}
\label{contact search condition}
\end{equation}
% where $\left<\Vec{a},\Vec{b}\right>$ denotes the function about computing the angle between two vectors $(\Vec{a},\Vec{b})$. 
% \textcolor{magenta}{if this function does the same thing as \eqref{arccosfunc}, please consider to unify.}

For a contact $c_k$, we consider it as qualified only if all the benchmarks $\{B_{kb}|b=1,2,3\}$ are satisfied based on the metrics.
% it we are capable to search the points meeting the conditions respect to the benchmarks $\{B_{kb}|b=1,2,3\}$ respectively, this contact $c_k$ is qualified. 
According to the benchmark set $\mathcal{B}$, we search the target contact along the sequence of $k=2,\cdots,q,1$, meaning that the first contact on the end is skipped and placed in the last of the queue. This priority distribution avoids breaking out the constructed contacts on the end while manipulating the central part of the DLO. Note that this search paradigm stops once the target contact $c_k$ is acquired along the defined sequence.
Our low-level planner takes the target contact $c_k$ as input and output the motion of the dual-arm robot $(\mathcal{T}_L,\mathcal{T}_R)$ to achieve it. The benchmark set $\mathcal{B}$ also serves as the guidance for contact construction, whereas there exists a gap between the keypoints $P^{[t]}$ and $\mathcal{B}$. 
% For grasping position planning, we focus on the keypoints $P^{[t]}$ instead of the raw state $S^{[t]}$ to reduce the search space. However, there exists a gap between the keypoints $P^{[t]}$ and the benchmarks $\mathcal{B}$.
To link the benchmark $\mathcal{B}$ and the keypoints $P^*$, we pair them up with the nearest Euclidean distance, using a set $\mathcal{J}=\{J_{kb}|k=1,\cdots,q,b=1,2,3\}$ to mark the corresponding index:
\begin{equation}
    J_{kb}=\arg\min_i||\Vec{p}_i^*-\Vec{B}_{kb}||_2
\end{equation}
With the set $\mathcal{J}$, the search sequence of grasping points selection for individual robots is divided into two cases 1) starting from the ends to the middle for $c_k$ on the end and 2) starting from the middle to the ends for $c_k$ in the intermediate.  
% $1\rightarrow J_{kb}, m\rightarrow J_{kb}$ for $c_k$ on the end and $J_{kb}\rightarrow1, J_{kb}\rightarrow m$  
% \textcolor{blue}{what the difference between $1\rightarrow J_{kb}$ and $J_{kb}\rightarrow1$? Also, all b=1,2,3 hold for $J_{kb}\rightarrow1$?}
This solution enables to limit the impacts caused by the manipulation on unrelated sections of the DLO.
% a larger range of the DLO between robots to be manipulated.
% \textcolor{blue}{This part is confusing. Where does 16 come from? (The number of keypoints?) 
% \textcolor{blue}{
% What's the difference between end and intermediate?}
This search paradigm undertakes under the constraints of the system, including the operation range and the contact obstacles.  
Next, we build on potential field \cite{song2002potential} path planning to avoid the collision with the contacts, which provide a repulsion force to the robot. Specifically, we extend the benchmark set $\mathcal{B}$ to $\mathcal{B}'$ with a threshold $\tau_b$:
\begin{equation}
B_{kb}'=\Vec{c}_k+\tau_b\cdot\overrightarrow{c_k B_{kb}}/||\Vec{B}_{kb}-\Vec{c}_k||_2,
\end{equation}
% This extended benchmark $\mathcal{B}'$ is effective to avoid the crash with thte contact $c_k$. 
% \textcolor{blue}{the first summation term is $\Vec{c}_k$ or $\Vec{B}_{kb}$?}
% We re-order the benchmarks $B_{kb}'\in\mathcal{B}'$ along the sequential keypoints $P^*$. 
% For each extended benchmark $B_{kb}'$, we assign a nearest keypoint to it, namely $B_{kb}\rightarrow p^*_{kb}$ and record their corresponding index $i$. Thie set is represented as $\mathcal{J}_k=\{J_{kb}\}$.
% \begin{equation}
%     J_{kb}=\arg\min_i||\Vec{p}^*_{i}-\Vec{B}'_{kb}||_2
% \end{equation}
% This extended benchmarks are guidelines to construct the contact. 
For the target contact $c_k$, we manipulate the DLO to the corresponding extended benchmark $\{B'_{kb}|b=1,2,3\}$. During the action, dual arms are assigned as fixing for freedom constraints and moving for shaping, depending on the reachability analysis of the robot arm concerning $B'_{k1}$. After that, the motion path is $B_{k3}$ to $B_{k_1}$ for left arm or $B_{k1}$ or $B_{k_3}$ for the right arm.

The 4-DOF pose on a table-top environment is defined as $\pi_j=\{\Vec{\chi}_j,\Vec{\eta}_j\}$, where $\Vec{\chi}_j$ and $\Vec{\eta}_j$ are position and direction vectors of $3\times1$. These two entities are defined by:
\begin{equation}
\Vec{\chi}_j=B'_{kb},\ \Vec{\eta}_j\cdot\overrightarrow{c_k B'_{kb}}=0
\end{equation}

At last, we give a summary about the primitive. With the target contact $c_k$, we select the grasping points for dual arms. Next, we assign their roles and execute the motion. At last, the robots release the DLO, waiting for the re-planning.

\subsection{Shape Primitive}
\begin{figure}
    \centerline{\includegraphics[width=\columnwidth]{"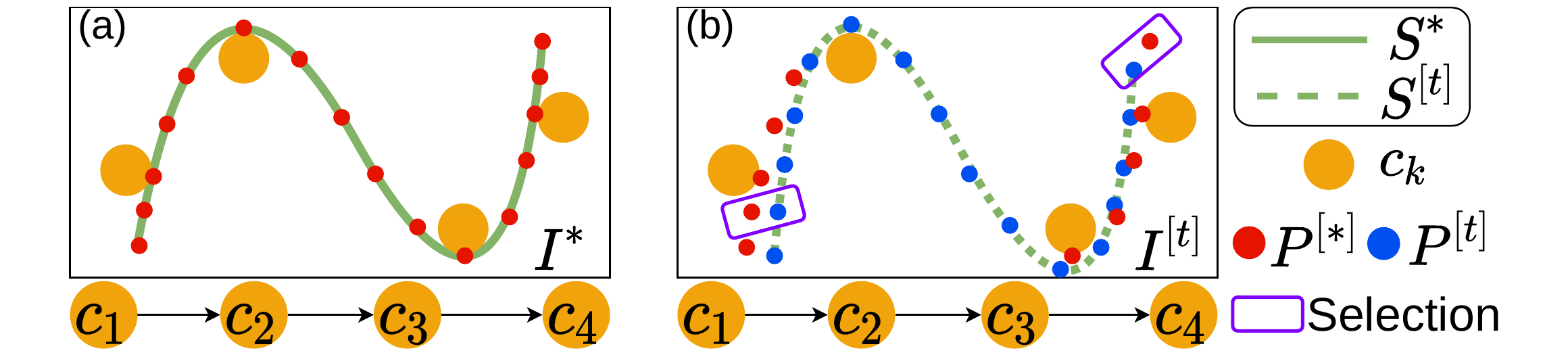"}}
    \caption{Graphical explanation of the shape primitive. (a) The desired goal guides the framework for shaping. (b) Based on the goal and the current state, we select the points of interest to reshape.
    % \textcolor{blue}{Is it possible to make the legends of all the figures consistent? In this figure we have the green line as S rather than P as marked previously.}
    } 
    \label{3.4-shape_primitive}
\end{figure}
The goal of the shape primitive is to implement the finetuning after the contact construction. 
% Similarly, we base on the detected keypoints $P^{[t]}$ and $P^*$ for hierarchical planning.
% to reduce the shape error during each action at time-step $t$.% Due to the state $S^{[t]}$ represented in the high-dimensional space, we shape the DLO through our detected keypoints $P^{[t]}$ and $P^*$. 
% arrange the configuration of DLOs precisely to minimize the error. 
As illustrated in Fig. \ref{3.4-shape_primitive}, our perception network correspondingly pairs up individual keypoints $p_j^{[t]}\in P^{[t]}$ and $p_j^*\in P^*$. Hence, the shape error $\Delta P$ between them is defined as:
\begin{equation}
    \Delta P=\frac{1}{m}\sum_{j=1}^{m}||p^{[t]}_j-p^*_j||_2
\end{equation}
This error serves as a guidance to improve the similarity. Intuitively, we select the keypoints $p_j$ whose difference between the current stage $P^{[t]}$ and the goal stage $P^{*}$ is comparatively large. Meanwhile, we also find the one in second-level for bimanual manipulation:
\begin{equation}
g\leftarrow\arg\max_j||p^{[t]}_j-p^*_j||_2,g'\leftarrow\arg\max_{j,j\neq g}||p^{[t]}_j-p^*_j||_2
\end{equation}
% \textcolor{blue}{$T_1$ is not defined.}
% According to our problem formulation, 
We reorder $(g,g')$ and reassign it to the dual-arm robot by
\begin{equation}
    g_L,g_R\leftarrow\min(g,g'),\max(g,g')
\end{equation} 
% After this target point selection, we apply the similar search paradigm in contact primitive. 
% Note that $g$ is used here to demonstrate the index of the selected keypoints. 
Similar to the contact primitive, we define the search paradigm under the system constraints as $g_L$ to $1$ for left arm and $g_R$ to $m$ for the right arm, respectively. Then, we define the target pose with respect to the g-th keypoint $p^*_g$:
\begin{equation}
    \pi_g=\{\Vec{p}_g^*,\delta p^*_g\}
\end{equation}
where $\delta p^*_g$ is tangent of $p^*_g$. 
% , defined as $\bigtriangledown p^*_i=\Vec{p^*_{i+1}}-\Vec{p^*_{i-1}}$. 
% \textcolor{magenta}{$G_R$ and $G_L$ are not defined.}
This shape primitive iterates until the desired goal is reached.
% The idea central to our shape finetuning (Alg. \ref{PlanShapeMotion}) is that reduce the biggest error within the keypoint $P^{[t]}$. 
% \begin{algorithm}
% 	\caption{PlanShapeMotion($P^{[t]},P^*$)}
% 	\label{PlanShapeMotion}
% 	\begin{algorithmic}[1]
% 	\STATE $T_1\leftarrow\arg\max_i||p^{[t]}_i-p^*_i||_2$
% 	\STATE $T_2\leftarrow\arg\max_{i,i\neq T_1}||p^{[t]}_i-p^*_i||_2$
% 	\STATE $T_L,T_R\leftarrow\min(T_1,T_2),\max(T_1,T_2)$
% 	\FOR{$p^{[t]}_i\in p^{[t]},i=T_L,T_L-1,\cdots$}
%         \IF{LeftRobotReachability($p_i^{[t]},\mathbb{T}^L$)}
%             \STATE $\Vec{\chi}_L\leftarrow$ PlanPose($P^{[t]},p^{[t]}_i$) \&\& Break
%         \ENDIF
% 	\ENDFOR
% 	\FOR{$p^{[t]}_i\in p^{[t]},i=T_R,T_R+1,\cdots$}
%         \IF{RightRobotReachability($p_i^{[t]},\mathbb{T}^R$)}
%             \STATE $\Vec{\chi}_R\leftarrow$ PlanPose($P^{[t]},p^{[t]}_i$) \&\& Break
%         \ENDIF
% 	\ENDFOR
% 	\RETURN $\Vec{\chi}_L,\Vec{\chi}_R$
% 	\end{algorithmic} 
% \end{algorithm}

% With the search index range ($T_L,T_R$), dual arms search the keypoints from its corresponding range to the corner. Similar to the above section, this search should satisfy the reachability condition of the robot. Different from the coarse contact mode, here we use keypoints of desired goal $k^*$ the for the finetuning:
% \begin{equation}
%     \pi_j=\{p_j^*,\overrightarrow{p^*_{j-1}p^*_{j+1}}\}
% \end{equation}
% This mode iterative until the error is smaller than the predefined threshold.

\section{RESULTS}
\subsection{Hardware Setup}
As illustrated in Fig. \ref{1-overview}, our bimanual experimental platform consists of two UR3 robotic manipulators, equipped with 2-fingered Robotiq grippers, respectively. To facilitate the bimanual manipulation, they face each other with an interval of $0.6m$.
An Intel Realsense L515 camera is mounted to sense the top-down view of the manipulation space with a resolution of $1280\times780$. The spatial transformation between the depth camera and dual-arms $(\mathbb{T}^L,\mathbb{T}^R)$ is calibrated through the markers. Each contact is a cylinder (radius=4cm,height=1cm), localized via ArUco markers. All contacts are glued on the table, keeping them stable during the whole manipulation process. The contacts are conventionally ordered according to the detected sequential keypoints $P^*$ concerning the goal shape of DLO $S^*$. Considering the physical limitations, the operation space of individual robots is constrained to a ring-shaped region. 

\subsection{Perception}
For perception in real environment, we utilize OpenCV \cite{bradski2000opencv} to segment the DLO $S^{[t]}$ from the raw observation $I^{[t]}$ with a morphological operation-based color filter, represented as a binary image. To balance the accuracy and efficiency, we resized $S^{[t]}$ to $128\times64$ for the following processing.
\begin{figure}
    \centerline{\includegraphics[width=\columnwidth]{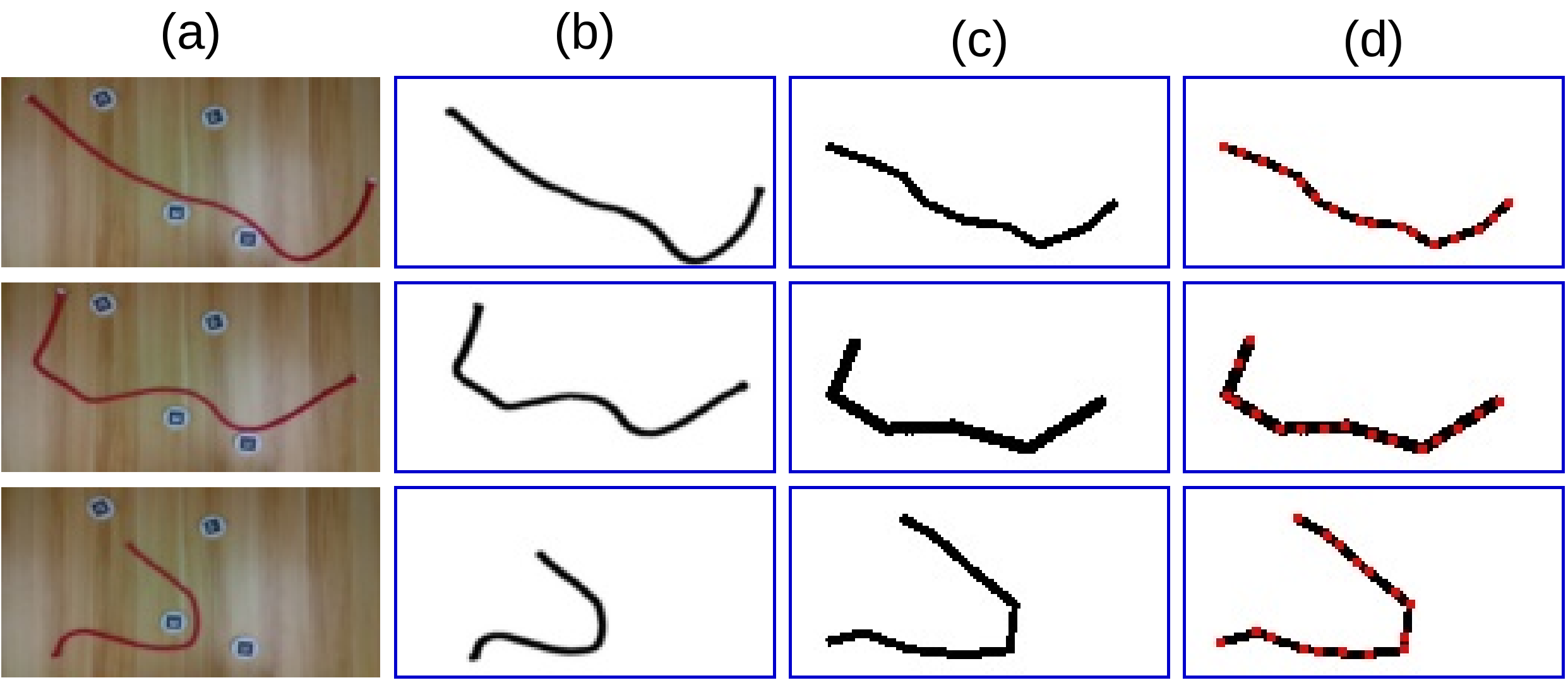}}
    \caption{Visualizations of synthetic dataset and the comparison with the real collected data. (a) Visual observation. (b) Extracted state of the DLO by the color filter. (c) Rendered state of the DLO. (d) Rendered keypoints of the DLO.}
    \label{4.1-dataset}
\end{figure}

In this section, we introduce the superiority of our synthetic-based feature extraction without any manual data collection and annotations.
To reduce the gap between simulation and reality, the synthetic dataset needs to render the physics. We quantitatively and qualitatively evaluate the robustness and accuracy of the perception model.

Fig. \ref{4.1-dataset} visualizes the synthetic dataset concerning the real data. Note that Fig. \ref{4.1-dataset}(a)-(b) is designed manually to act as references to have an intuitive comparison with the simulated Fig. \ref{4.1-dataset}(c)-(d). These graphical results validate the visual similarity with the real dataset.
% generated automatically with our rendering method, while Fig. \ref{4.1-dataset} (a)-(b) is manually designed configuration referring to them. 
% Intuitively, our simulated data $\hat{S}$ shares similar visual appearance with the extracted DLO $S$ from the environment $E$. 
% We also show the annotated keypoints of the synthetic image $\hat{S}$.
% Feature extraction is the prerequisite of the whole task. Compared with rigid objects, feature extraction in terms of deformable objects should trade off the representation level and the dimension.
% ranges from extracting key notes from the raw input (state estimation) and keep the balance between the accuracy and the dimension (dimension reduction). For our data-driven state estimation model, we provide some visualization results about our synthetic images dataset and statistical results about our representation level of our state estimator based on it.  
Our synthetic dataset includes 7040 labeled images in total, divided into a training dataset and testing dataset with a ratio of 10:1. Each sample is rendered as a binary image, containing a randomly generated curve and $m=16$ corresponding sorted keypoints in image coordinates. 
% are sampled based on the mathematical generation model as groundFully convolution network truth annotation of DLO states. 
To improve the variation of the dataset, the geometry features of the DLO, including radius, length, and the number of segments, are randomly generated over a wide range.
% during each generation. Fig. \ref{4.1-dataset} visualizes our synthetic dataset and the comparison with the real scenerio. Graphically, our dataset is similar to the real one.

% \begin{figure}
%     \centerline{\includegraphics[width=\columnwidth]{"figures/4.2-FCN_network"}}
%     \caption{Detailed architecture of our keypoint detection network.}
%     \label{4.2-FCN_network}
% \end{figure}

\begin{figure}
    \centerline{\includegraphics[width=\columnwidth]{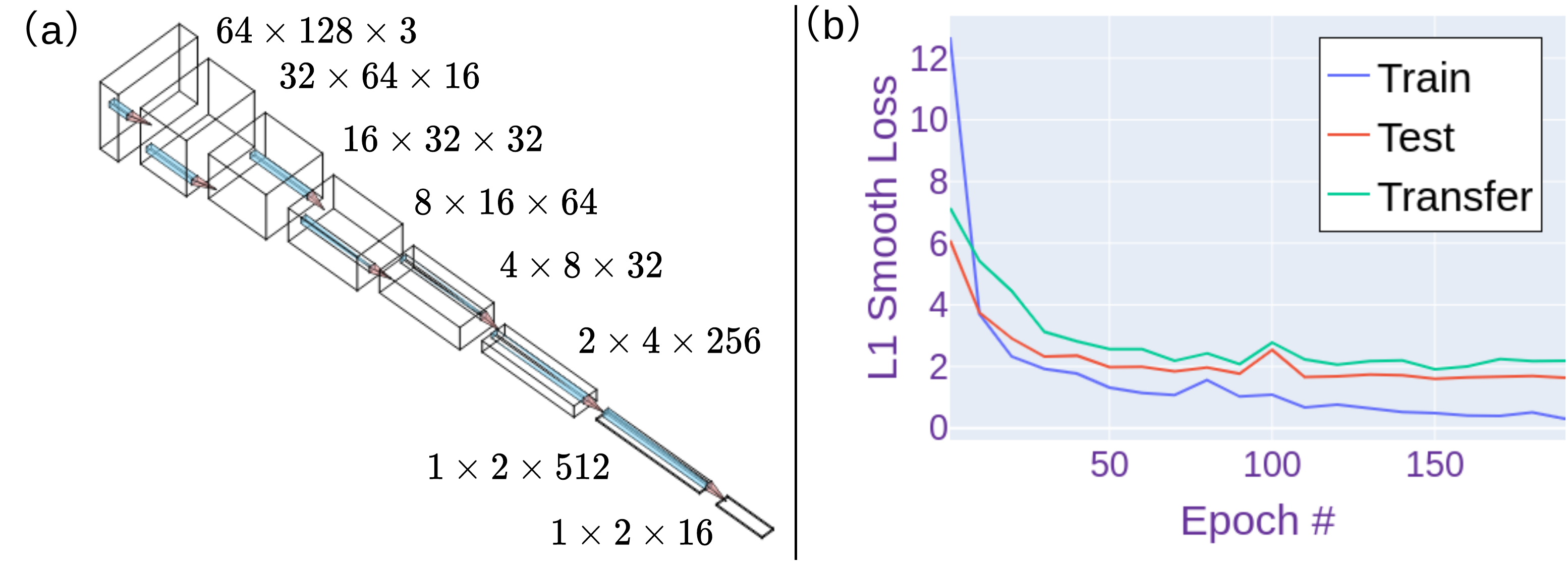}}
    \caption{Details about the perception network. (a) Architecture of the FCN network. (b) Loss convergence of training, validation, testing, and transferring.}
    \label{4.2-network_loss}
\end{figure}

Based on the synthetic dataset, we train our supervised keypoint detection network, whose architecture is shown in Fig. \ref{4.2-network_loss}(a). As a fully convolution network \cite{long2015fully}, it only involves convolution layers with a similar structure to VGG \cite{simonyan2014very}. In the last layer, we apply $1\times1$ convolution to regress the dimension of the output as $2\times16$, where each column represents the position $\overline{p}_j=(\overline{u}_j,\overline{v}_j)$ in the image frame. The training is optimized based on the smooth L1 loss function
\begin{equation}
\mathcal{L}_1(y,\hat{y})= \begin{cases}\frac{1}{2}(y-\hat{y})^{2} & \text { for }|y-\hat{y}| \leq 1 \\ |y-\hat{y}|-0.5 & \text { otherwise }\end{cases}
\end{equation}
where $y$ and $\hat{y}$ denote the ground truth and the output of the training, respectively.
Fig. \ref{4.2-network_loss}(b) shows the corresponding loss trend for training, testing, and transferring. Note that both training and testing are implemented with our synthetic dataset for efficient processing.
% We train our network on the synthetic dataset and evaluate its performance on the testing images, whose learning trend is shown in Fig. \ref{4.2-network_loss} (b). 
In addition, the transfer loss is evaluated on the real data collection with manual annotation, which includes fifty samples. Note that this manual collection dataset is only for evaluation and is not used to train the network. The promising results reveal the advantages of our perception method: 1) our synthetic dataset holds a high similarity with the real data to avoid manual collection; 2) the keypoint detection network converges to minimize the detection error; (3) the perception model is general to unseen samples in testing (simulation) and transferring (real).
\begin{figure}
    \centerline{\includegraphics[width=\columnwidth]{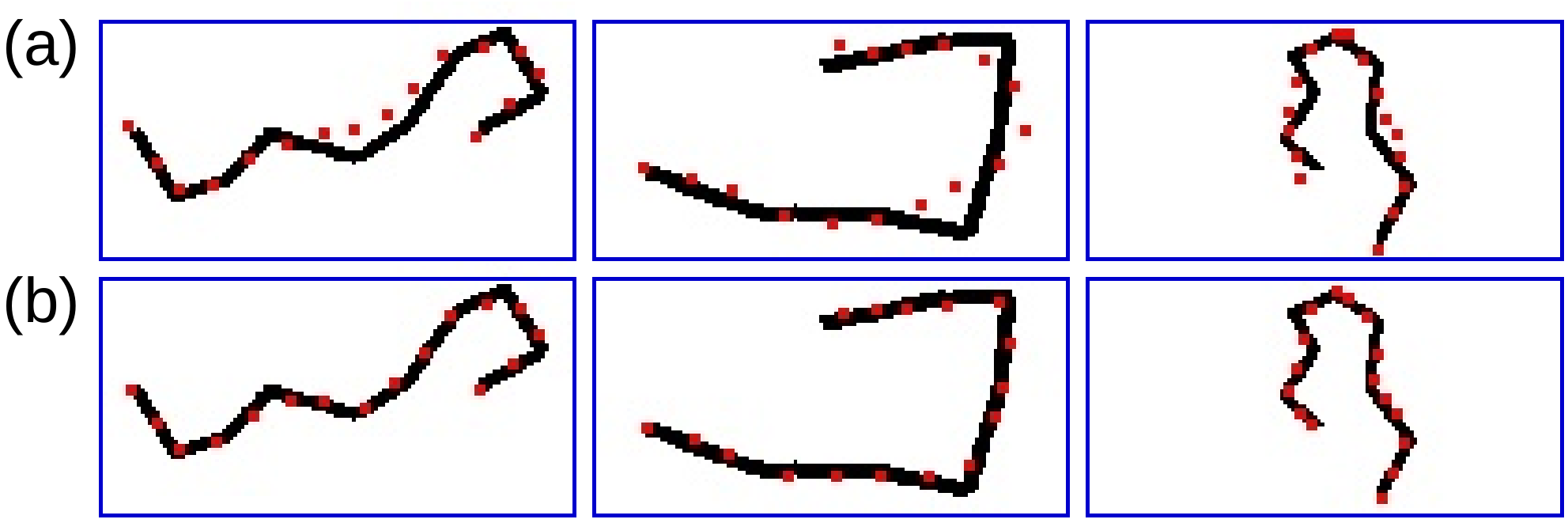}}
    \caption{Visualizations of finetuning the predicted keypoints according to the surrounding geometric features. (a) Raw output of the network. (b) Finetuning results.}
    \label{4.3-fine_tune}
\end{figure}

As discussed above, geometric finetuning is proposed to account for the residual error. Fig. \ref{4.3-fine_tune}(a) illustrates several failure cases, in which some detected keypoints drop out from the positive region of the DLO, mainly on the steep area of the curve. Comparatively, Fig. \ref{4.3-fine_tune}(b) visualizes the keypoints with finetuning, graphically indicating that this method improves the representation level of the sequential keypoints.

\begin{table}[!t]
\renewcommand{\arraystretch}{1}
\caption{COMPARISON OF KEYPOINT DETECTION PERFORMANCE}
\label{4.3-Keypoint}
\centering
\begin{tabular}{ccccc}
\toprule
 & \multicolumn{2}{c}{Corner Error $E_C$} & \multicolumn{2}{c}{Keypoint Error $E_P$}    \\
\hline
 & $\mu_C$ & $\sigma_C^2$ & $\mu_P$ & $\sigma_P^2$ \\
\midrule
Geo & 1.96 & 29.78 & 28.21 & 389.96 \\
Ours & 1.71 & 12.5 & 3.36 & 21.44 \\
\bottomrule
\end{tabular}\\
\begin{tablenotes}
\item Geo: Geometric-based method; Our: Our data-driven algorithm. $\mu_C,\sigma_C^2:$ Mean and variance of corner error $E_C$. $\mu_P,\sigma_P^2:$ Mean and variance of keypointd error $E_P$.
% $\sigma_{I}^2:$ Intra-class variance of region $S_j$. $\sigma_{B}^2:$ Between-class variance of region $S_j$. 
\end{tablenotes}
\end{table}

% Since our hierarchical planning framework depends on a sequence of ordered keypoints, their detection is crucial for the manipulation.
Compared with data-driven learning models, manual designed descriptor is an alternative for keypoint detection due to its intuitiveness and interpretability. Here, we provide a comparison between our method and a traditional geometric-based baseline, whose steps include skeletonizing DLOs via \cite{zhang1984fast} from $S^{[t]}$, searching the corners of DLOs according to the mesh grids, sorting and sampling the keypoints based on nearest neighbor search. Our error metrics include the corner $E_C$ and the keypoint detection error $E_P$, which are defined as $E_C=\frac{1}{2}(||\hat{p}_1-p_1||_2+||\hat{p}_m-p_m||_2)$ and $E_P=\frac{1}{m}\sum_{j=1}^{j=m}||\hat{p}_j-p_j||_2$, respectively. We emphasize the corner error $E_C$ here since it is the symbol to order the keypoints. Statistically, we leverage the mean value ($\mu_C,\mu_P$) and the variance ($\sigma_C^2,\sigma_P^2$) to evaluate their performance comprehensively. Note that $p_j$ and $\hat{p}_j$ are the ground truth of the dataset and the output of the corresponding algorithm, respectively. The comparison results are shown in Table. \ref{4.3-Keypoint}.
% For the evaluation of ordering sequence, we validate it through the similarity comparison between $\vec{\hat{p}}_i$ and $\vec{p}_i$. 
Due to the huge diversity of the state space of DLOs, it is very difficult to manually develop a sequential keypoint detection method that is robust to various configurations. Conversely, our perception network is robust with its data-driven manner.

% \begin{table}[!t]
% % increase table row spacing, adjust to taste
% \renewcommand{\arraystretch}{1}
% \caption{NETWORK ARCHITECTURE OF THE COMPARISON BASELINES}
% \label{4.1-Network Structure of comparison}
% \centering
% \begin{tabular}{cccc}
% \toprule
% \textbf{N} & LR ($1\times8192$) & CNN ($64\times128$) & PC ($50\times3$) \\
% \midrule
% E1 & FC(8192,2048), R & Cv(3,3), 16, R, P4 & Cv(1,1), 8, BN, R \\
% E2 & FC(2048,256), R & Cv(3,3), 4, R, P2 & Cv(1,1), 16, BN, R  \\
% E3 & FC(256,32), R & Cv(3,3), 1, R, P2 & Cv(1,1), 32, BN, R  \\
% L & $1\times32$ & $4\times8\times1$ & $1\times32$  \\
% D1 & FC(32,256), R & I4, Cv(3,3), 16, R & FC(32,64), R  \\
% D2 & FC(256,2048), R & I2, Cv(3,3), 4, R & FC(64,128), R  \\
% D3 & FC(2048,8192), S & I2, Cv(3,3), 1, S & FC(128,150)  \\
% $\mathcal{L}$ & BCE & BCE & Chamfer\\
% \bottomrule
% \end{tabular}\\
% \begin{tablenotes}
% \item $E$-Encoder, $D$-Decoder, $\mathcal{L}$ is loss function.
% % $d_{min}$ and $d_{max}$ is minimum and maximum value of the Euclidean distance. $\mu_{D}$ and $\sigma_D$ is the mean and standard variance of the distance respectively.
% \end{tablenotes}
% \end{table}
A key issue about descriptors is their representation level versus the original data.
% We evaluate the representation level of our state estimation on simulated data.
% One of the most important criteria of state estimation is its representation level of the original data. In this section, we prove the advantages of our estimator through the reconstruction error. 
Since we only predict keypoints of DLOs based on the link-chain model, we reconstruct the original shape through end-to-end connection. For comparisons, we consider various unsupervised auto-encoders \cite{zhou2021lasesom}, whose goal is also to extract a compact latent code about the high-dimensional data. We choose three baselines to adapt to our case 1) fully connected linear regression (LR), 2) convolutional neural network (CNN), and 3) PointNet \cite{achlioptas2018learning} (PC). 
% , latent space approaches have recently achieved many successful results, due to its capability to encode high-dimensional unorder raw data to a compact vector in latent space. As a result, we have a comparison between our network the related latent-based autoencoders. The details of the networks are illustrated in \ref{4.1-Network Structure of comparison}. 
Specifically, the training of LR and CNN autoencoders is conducted based on the binary cross entropy (BCE) loss $\mathcal{L}_{BCE}$, while PC autoencoder is optimized through Chamfer distance $d$. They are defined as: 
% During training, the selected BCE loss and chamfer loss are:
\begin{equation}
\begin{aligned}
&\mathcal{L}_{BCE}=-\sum_{i=1}^{n} y_i \log \hat{y}_i+\left(1-y_i\right) \log \left(1-\hat{y}_i\right)\\
&d\left(\hat{Y},Y\right)=\sum_{\hat{y}\in\hat{Y}} \min _{y \in Y}\|\hat{y}-y\|_{2}^{2}+\sum_{y \in Y} \min _{\hat{y} \in \hat{Y}}\|\hat{y}-y\|_{2}^{2}   
\end{aligned}
\end{equation}
According to the network structure, LR and CNN take the 2D image format as input while PC utilizes the 3D point cloud with the same size after down-sampling.

% \begin{table}[!t]
% % increase table row spacing, adjust to taste
% \renewcommand{\arraystretch}{1}
% \caption{Pixel comparison}
% \label{2-REGISTRATION ERROR}
% \centering
% \begin{tabular}{cccccc}
% \toprule
% \textbf{Estimator} & Pixel & BCE & $A\cap B$ & $A\cup B$ & IoU   \\
% \midrule
% FCN & 0.004 & 0.680 & 0.167 & 1.134 & 1.123 \\
% FCN2 & 0.004 & 0.680 & 0.167 & 1.134 & 1.123 \\
% FC & 0.004 & 0.680 & 0.167 & 1.134 & 1.123 \\
% CNN & 0.004 & 0.680 & 0.167 & 1.134 & 1.123 \\
% PC & 0.004 & 0.680 & 0.167 & 1.134 & 1.123 \\
% \bottomrule
% \end{tabular}\\
% \begin{tablenotes}
% \item $d_{min}$ and $d_{max}$ is minimum and maximum value of the Euclidean distance. $\mu_{D}$ and $\sigma_D$ is the mean and standard variance of the distance respectively.
% \end{tablenotes}
% \end{table}

\begin{table}
% [!htbp]
\centering
\caption{COMPARISON OF KEYPOINT DETECTION PERFORMANCE ON SYNTHETIC DATASET}
\label{4.2-Reconstruction}
\begin{tabular}{|c|c|c|c|c|c|c|}
%\begin{tabularx}{\textwidth}{|c|c|c|c|c|}
\hline
 & \multicolumn{3}{c|}{L1} & \multicolumn{3}{c|}{IoU}   \\
\hline
\textbf{Net} & Train & Valid & Test & Train & Valid & Test \\
\hline
FCN-L & 0.0074 & 0.0074 & 0.0073 & 0.6645 & 0.665 & 0.6648\\
\hline
FCN-R & 0.0274 & 0.0276 & 0.0272 & 0.0798 & 0.0783 & 0.0804\\
\hline
FCN-F & 0.0208 & 0.0212 & 0.0205 & 0.2558 & 0.2493 & 0.2573\\
\hline
LR & 0.0297 & 0.0299 & 0.0325 & 0.0846 & 0.0852 & 0.0643\\
\hline
CNN & 0.0294 & 0.0296 & 0.0291 & 0.0996 & 0.0995 & 0.0991\\
\hline
PC \cite{achlioptas2018learning} & 0.02 & 0.0201 & 0.0199 & 0.0882 & 0.0878 & 0.0852\\
\hline
\end{tabular}\\
\begin{tablenotes}
\item FCN-L: label of the FCN; FCN-R: raw output of the FCN; FCN-F: finetuning FCN; LR: linear regression; CNN: convolutional neural network; PC \cite{achlioptas2018learning}: point cloud. 
\end{tablenotes}
\end{table}

Since our original state $S^{[t]}$ is a binary image, the shape reconstruction issue here is formulated as a classification concerning each pixel $S^{[t]}(u,v)$. Hence, our evaluation metrics are L1 loss $\mathcal{L}_{1}$ and IoU (Intersection over Union) between the reconstructed output and the original information, respectively:
% \begin{equation}
%     \ell(x, y)=L=\left\{l_{1}, \ldots, l_{N}\right\}^{\top}, \quad l_{n}=\left|x_{n}-y_{n}\right|
% \end{equation}
\begin{equation}
\label{iou}
\mathcal{L}_{1}=\sum_{i=1}^n|y_i-\hat{y}_i|,
IoU=\frac{\hat{y}\cap y}{\hat{y}\cup y}
\end{equation}
Table. \ref{4.2-Reconstruction} shows the comparison results. Note that FCN-L method utilizes the labeled keypoints for reconstruction and acts as ground truth for our data-driven representation. The finetuning output of our perception improves greatly compared with the raw output of the network FCN-R. Compared with LR and CNN autoencoders, our proposed FCN-F performs better both in L1 loss and IoU. The main reason is that autoencoders aim to reconstruct the entire information of the input (even the details) instead of paying attention to the fundamental features. Although \cite{achlioptas2018learning} achieves well in L1 loss, its performance concerning IoU is poor. This is because it is only able to reconstruct the original data with a fixed size (due to the identical input dimension); thus loses some information inevitably. 

\subsection{Manipulation}
\begin{figure}
    \centerline{\includegraphics[width=\columnwidth]{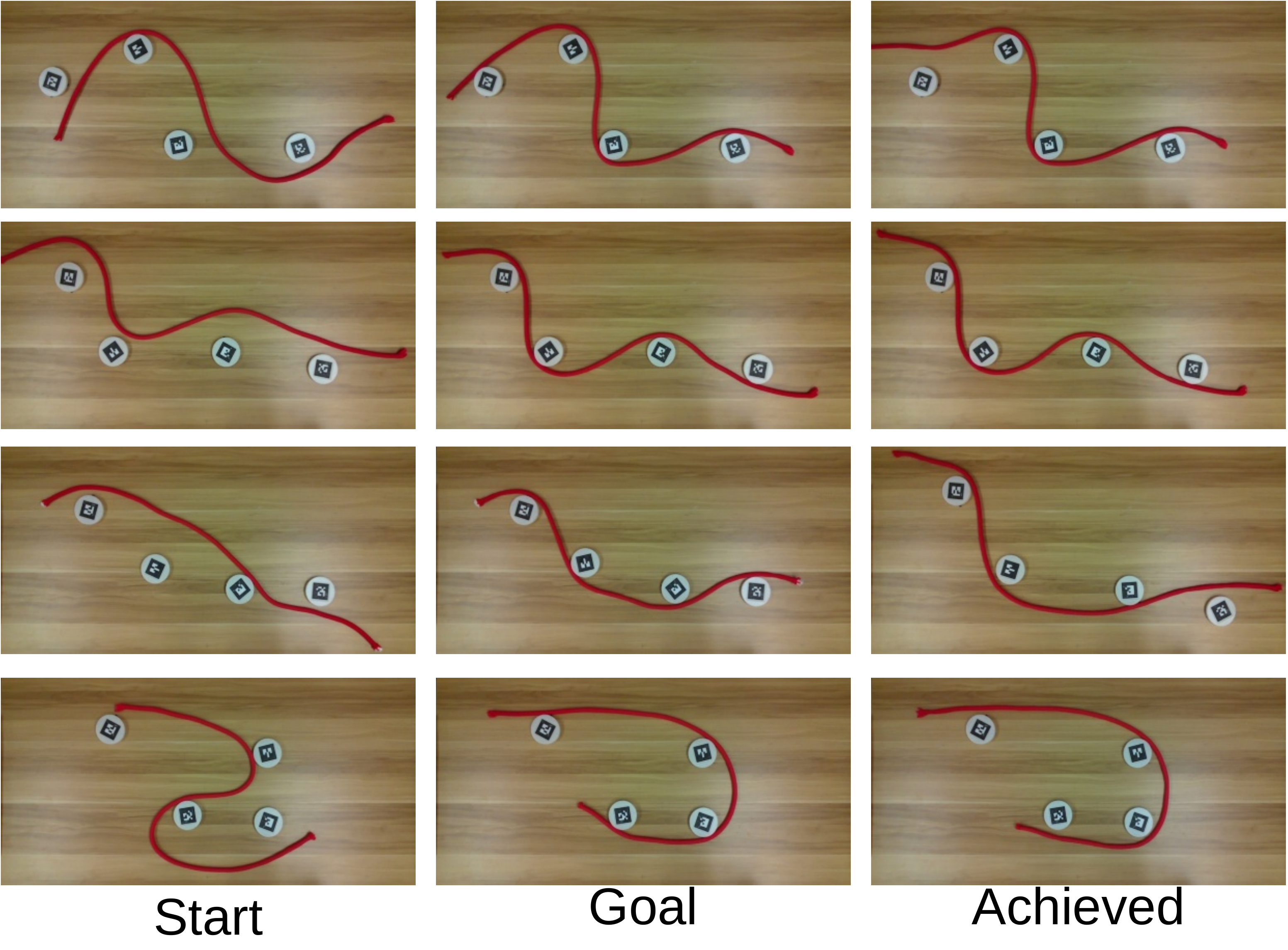}}
    \caption{Our designed DLO manipulation with environmental contacts scenarios. From left to right: the start state, the goal state and achieved state with our framework. All the images are taken by our top-down Realsense L515 depth camera.}
    \label{4.4-mani exp}
\end{figure}
To validate our hierarchical action planning framework, we evaluate the performance with multiple experiments using various contact configurations and goals. Fig. \ref{4.4-mani exp} shows four designed tasks in our experiment. 
% At the beginning of each task, a goal image $I^*$ is provided to guide our manipulation. We compute several key information from $I^*$: 1) segment the DLO $S^*$ with color filter and our perception network outputs corresponding $P^*$ and 2) localize the contacts $C=\{c_1,\cdots,c_k,\cdots,c_q\}$ and compute the contact-based benchmarks $(\mathcal{B},\mathcal{B}',\mathcal{J})$. 
Note that the configuration of the DLO at the beginning $S^{[0]}$ is placed randomly on the table and the desired goal is provided artificially. For each experiment, we assume that the goal shape $S^*$ keeps stable with the support of the contacts and the table. The third column in Fig. \ref{4.4-mani exp} illustrates our achieved results. Since our hierarchical action planning is iterative, the robot continuously manipulates the DLO until the shape similarity between the goal $S^*$ and the achieved one $S^{[H]}$ is sufficient. In this experimental study, the shaping tasks are conducted with multi-step action depending on the feature extraction of the DLOs without learning their physical dynamics.
\begin{figure}
    \centerline{\includegraphics[width=\columnwidth]{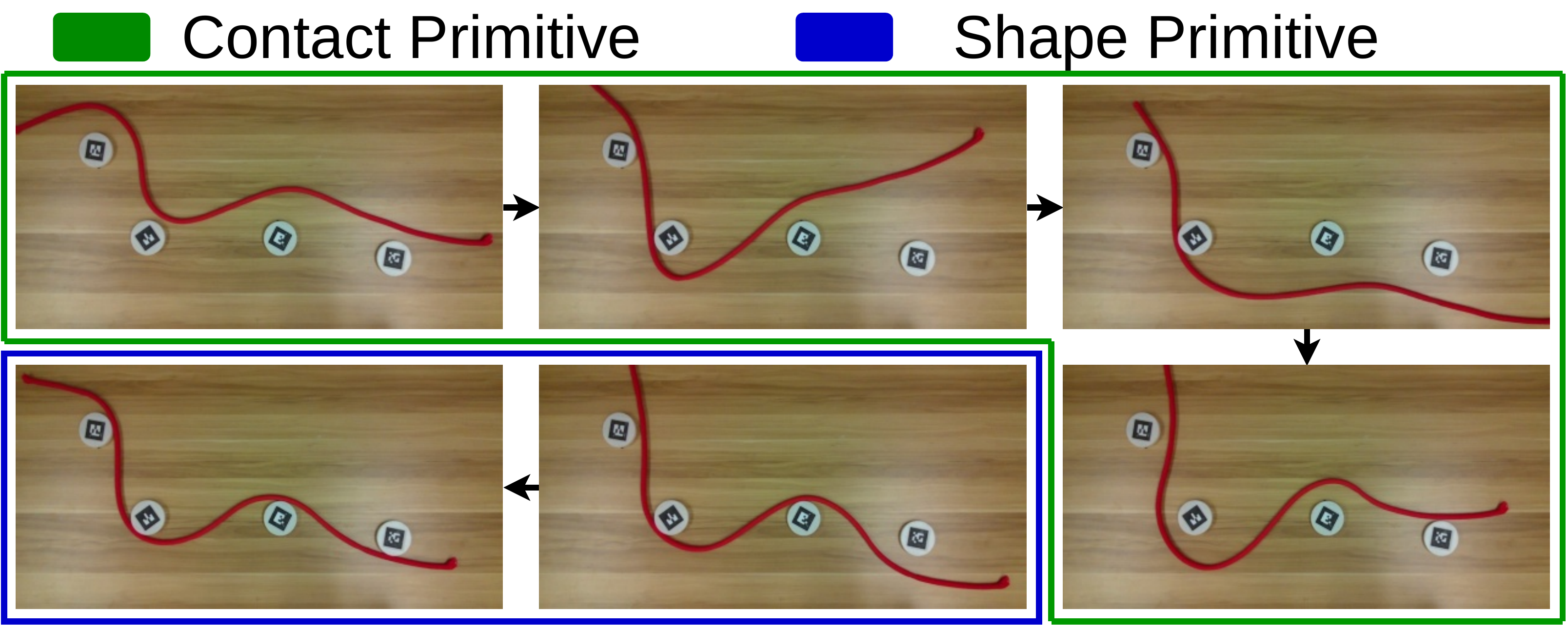}}
    \caption{Our designed DLOs manipulation with environmental contacts scenarios. From left to right: the start state, the goal state and achieved state with our framework. All the images are taken by our top-down Realsense L515 depth camera. }
    \label{4.5-mani_scene}
\end{figure}

As a multi-step decision-making process, we provide a typical example of the manipulation, as shown in Fig. \ref{4.5-mani_scene}. At the beginning, our algorithm computes the prior knowledge for the hierarchical action planning based on the goal image $I^*$: 1) segment the DLO $S^*$ with the color filter and detect the corresponding sequence ordered keypoints $P^*$ through our perception network and 2) localize the contacts $C=\{c_1,\cdots,c_k,\cdots,c_q\}$ and compute the contact-based benchmarks $(\mathcal{B},\mathcal{B}',\mathcal{J})$. 
Then, our algorithm enters into the action loop. For each planning, we sense the DLO $S^{[t]}$ and detect its keypoints $P^{[t]}$ via our perception network. With this, we check the contact construction based on our search benchmarks $\mathcal{B}$. If it is failed, we utilize the contact primitive to construct the corresponding contact $c_k$. Once the action plan ($\mathcal{T}_L,\mathcal{T}_R$) is accomplished, we update the state of the DLO $S^{[t+1]}$. If the contact restrictions are met, we move on to the shape primitive for finetuning. The entire algorithm iterates until reaching the goal state $S^*$, which the criteria is defined as the binary IoU between $S^{[t]}$ and $S^*$ according to Eq. \ref{iou} should be larger than $40\%$. 
We also provide supplementary material for robotic bimanual manipulation videos. 

% depicts the initial and final configurations for these experiments. In this experimental study, the tasks are conducted without the identification of the object's physical dynamics. Fig. \ref{4.7-Mani_scene} (a) and (b) visually depict the manipulation process of contact primitive and shape primitive respectively. For contact primitive, we need to select the grasp point and then arrange the motion. The motion path includes a series of waypoints to implement bypassing. For shape primitive, we need to choose grasp points and move to target points. Fig. \ref{4.7-Mani_scene} (c) gives an example of the whole manipulation process. Starting from the arbitrary initial configuration, we implement two times of contact primitive to establish the contacts. Passing the contact check, we utilize the shape primitive to iteratively minimize the local shape error until meeting the criteria.

\begin{figure}
    \centerline{\includegraphics[width=\columnwidth]{"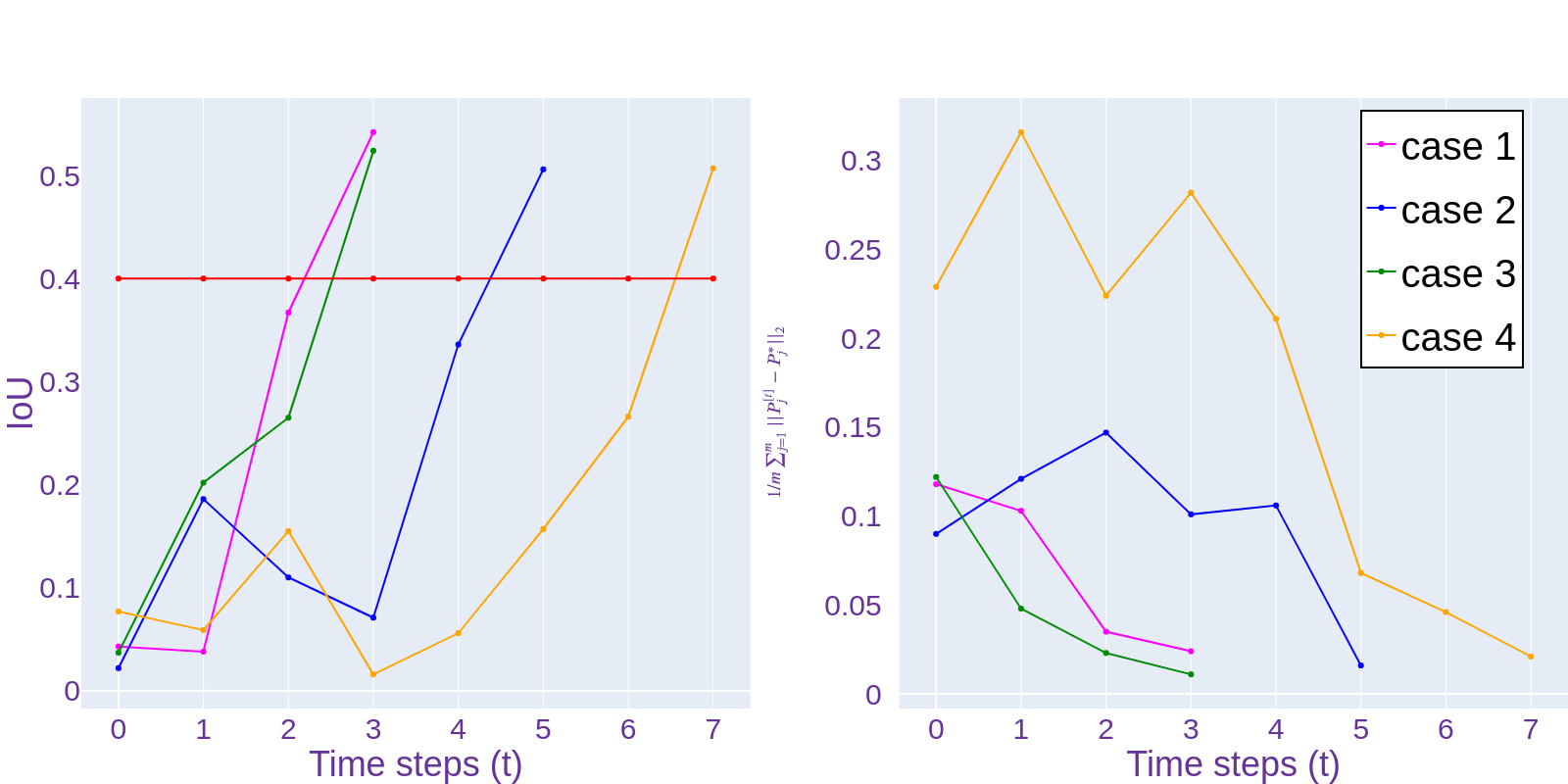"}}
    \caption{ Shape error minimization process. (a) IoU between the goal state $S^*$ and the state $S^{[t]}$ in each time step $t$. (b) The keypoints error between $P^*$ and $P^{[t]}$.}
    \label{4.6-shape_error}
\end{figure}
% Our method successfully finished all 4 tasks, visualized in Fig. \ref{4.5-mani_scene}. We also provide a supplementary material for robot bimanual manipulation videos. 

Based on the goal shape in Fig. \ref{4.4-mani exp}, we implement four trials under various initial configurations. Fig. \ref{4.6-shape_error} depicts the quantitative measurements of the scenarios in Fig. \ref{4.4-mani exp}. Specifically, the minimization of the magnitude error $\Delta P$ is shown in Fig. \ref{4.6-shape_error}(b). 
% We implement four trials under various initial configurations for each designed goal and only record the error after each action execution in discrete time. 
% The shape variations $||\Delta P||$ and IoU during the manipulation process for the scenerios in Fig. \ref{4.4-mani exp} are shown in Fig. \ref{4.6-shape_error}.
These results corroborate that the detected sequential keypoints can be used to manipulate the DLO into the desired specification. Fig. \ref{4.6-shape_error}(a) demonstrates the similarity level of the state at each time step with the goal shape $S^*$, which IoU$=40\%$ serves as a baseline. Note that the IoU value decreases compared with the previous time step in some cases since the contact-based manipulation task is not continuous. Hence, a coarse-to-fine manner is necessary for this challenging task, otherwise, we probably get stuck in a local optimum.
% our keypoint-based action planning for shaping DLOs is effective to achieve the goal. For a shape servoing task, the original data instead of our personal feature descriptors should be used to evaluate the performance. Here, we adopt the IoU between the goal state $S^*$ and the achieved state $S^{[H]}$, as shown in Fig. \ref{4.6-shape_error}(a). 
These results also reveal that our algorithm is superior in feature description and action planning versus this kind of challenging task.   
% The convergence conditions of the manipulation are set as $\Delta P<2cm$ and IoU$>40\%$ respectively. 
% These results corroborate that the keypoint-based task specification can be used to finish the deformable linear objects shaping tasks with environmental contacts. 

\begin{figure}
    \centerline{\includegraphics[width=\columnwidth]{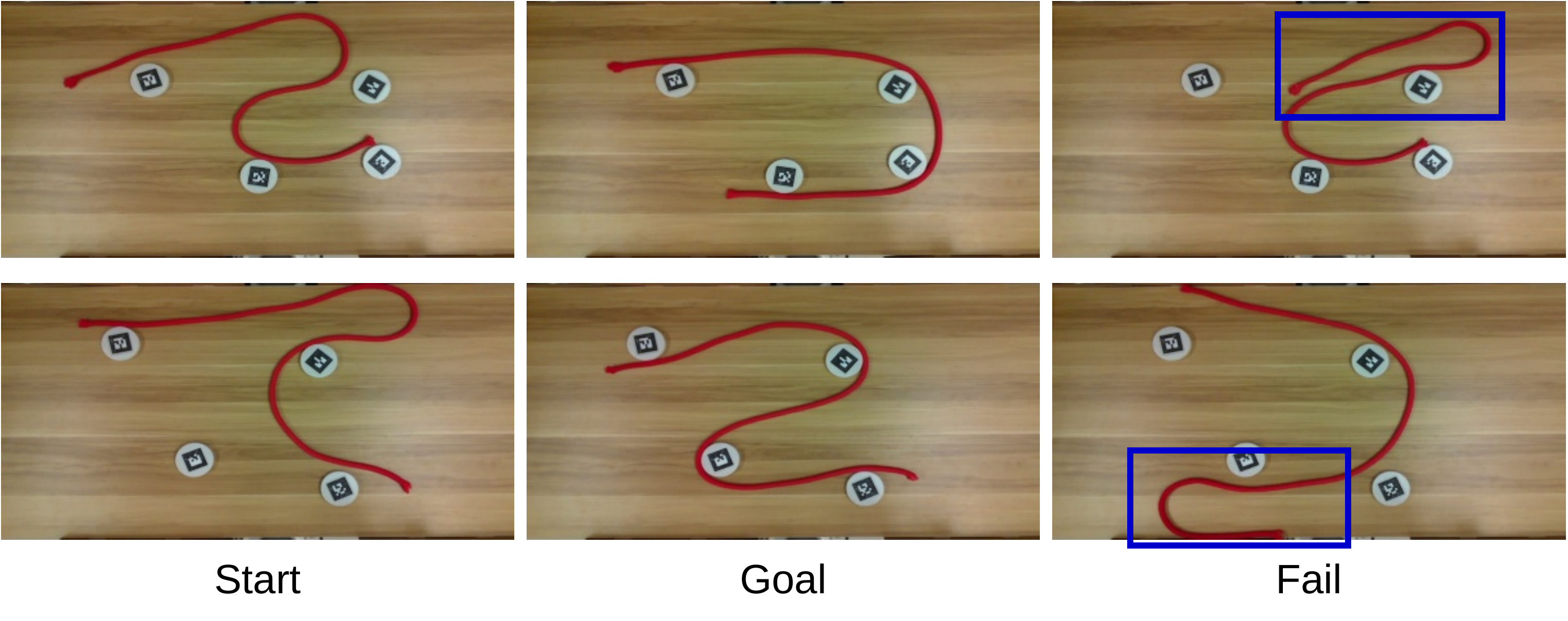}}
    \caption{The stuck state of failure cases. }
    \label{4.7-failed_case}
\end{figure}
Although our planning framework is capable of dealing with the majority of these challenging tasks, there are some cases that the system fails. Fig. \ref{4.7-failed_case} presents two typical failure examples. Although our perception network plays well in most cases, its performance is severely affected by rolling. That is because the convolution is not good at dealing with the details of the pixels and the finetuning regresses the keypoints to the wrong region of the DLO, resulting in a sequence of disordered keypoints. Another case is caused by the lack of physical dynamics. Without any forecasting and feedback, our framework replans the action in an open-loop form. Thus, the system probably enters into a local convergence at the contacts. 
% self-collision. To simplify the control strategy, we ignore the physical dynamics model and didn't provide some close-loop feedback during each action. 
% If the situation reaches out of the roadmap greatly, our strategy probably be not capable of restoring the common situation. 

\section{CONCLUSIONS}
In this paper, we demonstrate a keypoint-based bimanual manipulation for DLOs under environmental contact constraints. Training on a synthetic image dataset, our perception extracts sequential keypoints of DLOs as descriptors. The hierarchical action planning framework performs the task with two defined primitives in a coarse-to-fine manner. The whole algorithm is semantic without requiring any manual data collection and annotation. However, our methods exist some limitations. The perception network has poor performance in the knotted cases. As an open-loop method, the stability of the planner is not guaranteed. For future directions, we are interested to explore the synergistic behaviors between dual arms to extend the framework for other deformable objects, such as clothes and bags.
% the collaboration between dual arms can be strengthened to deal with more complex cases. We also plan to In addition, the manipulated object can be extended to other kinds of deformable objects, such as clothes and bags.
% \addtolength{\textheight}{-12cm}   

%%%%%%%%%%%%%%%%%%%%%%%%%%%%%%%%%%%%%%%%%%%%%%%%%%%%%%%%%%%%%%%%%%%%%%%%%%%%%%%%

%%%%%%%%%%%%%%%%%%%%%%%%%%%%%%%%%%%%%%%%%%%%%%%%%%%%%%%%%%%%%%%%%%%%%%%%%%%%%%%%

%%%%%%%%%%%%%%%%%%%%%%%%%%%%%%%%%%%%%%%%%%%%%%%%%%%%%%%%%%%%%%%%%%%%%%%%%%%%%%%%

%%%%%%%%%%%%%%%%%%%%%%%%%%%%%%%%%%%%%%%%%%%%%%%%%%%%%%%%%%%%%%%%%%%%%%%%%%%%%%%%

% \small
\bibliographystyle{ieeetr}
\bibliography{ref}

\begin{thebibliography}{10}

\bibitem{zhu2018dual}
J.~Zhu, B.~Navarro, P.~Fraisse, A.~Crosnier, and A.~Cherubini, ``Dual-arm
  robotic manipulation of flexible cables,'' in {\em 2018 IEEE/RSJ
  International Conference on Intelligent Robots and Systems (IROS)},
  pp.~479--484, IEEE, 2018.

\bibitem{garcia2020benchmarking}
I.~Garcia-Camacho, M.~Lippi, M.~C. Welle, H.~Yin, R.~Antonova, A.~Varava,
  J.~Borras, C.~Torras, A.~Marino, G.~Alenya, {\em et~al.}, ``Benchmarking
  bimanual cloth manipulation,'' {\em IEEE Robotics and Automation Letters},
  vol.~5, no.~2, pp.~1111--1118, 2020.

\bibitem{navarro2016automatic}
D.~Navarro-Alarcon, H.~M. Yip, Z.~Wang, Y.-H. Liu, F.~Zhong, T.~Zhang, and
  P.~Li, ``Automatic 3-d manipulation of soft objects by robotic arms with an
  adaptive deformation model,'' {\em IEEE Transactions on Robotics}, vol.~32,
  no.~2, pp.~429--441, 2016.

\bibitem{galassi2021robotic}
K.~Galassi and G.~Palli, ``Robotic wires manipulation for switchgear cabling
  and wiring harness manufacturing,'' in {\em 2021 4th IEEE International
  Conference on Industrial Cyber-Physical Systems (ICPS)}, pp.~531--536, IEEE,
  2021.

\bibitem{sanchez2018robotic}
J.~Sanchez, J.-A. Corrales, B.-C. Bouzgarrou, and Y.~Mezouar, ``Robotic
  manipulation and sensing of deformable objects in domestic and industrial
  applications: a survey,'' {\em The International Journal of Robotics
  Research}, vol.~37, no.~7, pp.~688--716, 2018.

\bibitem{navarro2013model}
D.~Navarro-Alarcon, Y.-H. Liu, J.~G. Romero, and P.~Li, ``Model-free visually
  servoed deformation control of elastic objects by robot manipulators,'' {\em
  IEEE Transactions on Robotics}, vol.~29, no.~6, pp.~1457--1468, 2013.

\bibitem{zhu2021vision}
J.~Zhu, D.~Navarro-Alarcon, R.~Passama, and A.~Cherubini, ``Vision-based
  manipulation of deformable and rigid objects using subspace projections of 2d
  contours,'' {\em Robotics and Autonomous Systems}, vol.~142, p.~103798, 2021.

\bibitem{navarro2017fourier}
D.~Navarro-Alarcon and Y.-H. Liu, ``Fourier-based shape servoing: a new
  feedback method to actively deform soft objects into desired 2-d image
  contours,'' {\em IEEE Transactions on Robotics}, vol.~34, no.~1,
  pp.~272--279, 2017.

\bibitem{zhu2021challenges}
J.~Zhu, A.~Cherubini, C.~Dune, D.~Navarro-Alarcon, F.~Alambeigi, D.~Berenson,
  F.~Ficuciello, K.~Harada, X.~Li, J.~Pan, {\em et~al.}, ``Challenges and
  outlook in robotic manipulation of deformable objects,'' {\em arXiv preprint
  arXiv:2105.01767}, 2021.

\bibitem{wang2018unified}
Z.~Wang, X.~Li, D.~Navarro-Alarcon, and Y.-h. Liu, ``A unified controller for
  region-reaching and deforming of soft objects,'' in {\em 2018 IEEE/RSJ
  International Conference on Intelligent Robots and Systems (IROS)},
  pp.~472--478, IEEE, 2018.

\bibitem{navarro2014visual}
D.~Navarro-Alarcon, Y.-h. Liu, J.~G. Romero, and P.~Li, ``On the visual
  deformation servoing of compliant objects: Uncalibrated control methods and
  experiments,'' {\em The International Journal of Robotics Research}, vol.~33,
  no.~11, pp.~1462--1480, 2014.

\bibitem{zhou2021lasesom}
P.~Zhou, J.~Zhu, S.~Huo, and D.~Navarro-Alarcon, ``Lasesom: A latent and
  semantic representation framework for soft object manipulation,'' {\em IEEE
  Robotics and Automation Letters}, 2021.

\bibitem{tang2018framework}
T.~Tang, C.~Wang, and M.~Tomizuka, ``A framework for manipulating deformable
  linear objects by coherent point drift,'' {\em IEEE Robotics and Automation
  Letters}, vol.~3, no.~4, pp.~3426--3433, 2018.

\bibitem{tanaka2018emd}
D.~Tanaka, S.~Arnold, and K.~Yamazaki, ``Emd net: An encode--manipulate--decode
  network for cloth manipulation,'' {\em IEEE Robotics and Automation Letters},
  vol.~3, no.~3, pp.~1771--1778, 2018.

\bibitem{johns2016deep}
E.~Johns, S.~Leutenegger, and A.~J. Davison, ``Deep learning a grasp function
  for grasping under gripper pose uncertainty,'' in {\em 2016 IEEE/RSJ
  International Conference on Intelligent Robots and Systems (IROS)},
  pp.~4461--4468, IEEE, 2016.

\bibitem{seita2020deep}
D.~Seita, A.~Ganapathi, R.~Hoque, M.~Hwang, E.~Cen, A.~K. Tanwani,
  A.~Balakrishna, B.~Thananjeyan, J.~Ichnowski, N.~Jamali, {\em et~al.}, ``Deep
  imitation learning of sequential fabric smoothing from an algorithmic
  supervisor,'' in {\em 2020 IEEE/RSJ International Conference on Intelligent
  Robots and Systems (IROS)}, pp.~9651--9658, IEEE, 2020.

\bibitem{sundaresan2020learning}
P.~Sundaresan, J.~Grannen, B.~Thananjeyan, A.~Balakrishna, M.~Laskey, K.~Stone,
  J.~E. Gonzalez, and K.~Goldberg, ``Learning rope manipulation policies using
  dense object descriptors trained on synthetic depth data,'' in {\em 2020 IEEE
  International Conference on Robotics and Automation (ICRA)}, pp.~9411--9418,
  IEEE, 2020.

\bibitem{yan2020self}
M.~Yan, Y.~Zhu, N.~Jin, and J.~Bohg, ``Self-supervised learning of state
  estimation for manipulating deformable linear objects,'' {\em IEEE robotics
  and automation letters}, vol.~5, no.~2, pp.~2372--2379, 2020.

\bibitem{jin2019robust}
S.~Jin, C.~Wang, and M.~Tomizuka, ``Robust deformation model approximation for
  robotic cable manipulation,'' in {\em 2019 IEEE/RSJ International Conference
  on Intelligent Robots and Systems (IROS)}, pp.~6586--6593, IEEE, 2019.

\bibitem{tang2018track}
T.~Tang and M.~Tomizuka, ``Track deformable objects from point clouds with
  structure preserved registration,'' {\em The International Journal of
  Robotics Research}, p.~0278364919841431, 2018.

\bibitem{yin2021modeling}
H.~Yin, A.~Varava, and D.~Kragic, ``Modeling, learning, perception, and control
  methods for deformable object manipulation,'' {\em Science Robotics}, vol.~6,
  no.~54, 2021.

\bibitem{5980391}
L.~P. Kaelbling and T.~Lozano-Pérez, ``Hierarchical task and motion planning
  in the now,'' in {\em 2011 IEEE International Conference on Robotics and
  Automation}, pp.~1470--1477, 2011.

\bibitem{wells2019learning}
A.~M. Wells, N.~T. Dantam, A.~Shrivastava, and L.~E. Kavraki, ``Learning
  feasibility for task and motion planning in tabletop environments,'' {\em
  IEEE robotics and automation letters}, vol.~4, no.~2, pp.~1255--1262, 2019.

\bibitem{zhu2019robotic}
J.~Zhu, B.~Navarro, R.~Passama, P.~Fraisse, A.~Crosnier, and A.~Cherubini,
  ``Robotic manipulation planning for shaping deformable linear objects
  withenvironmental contacts,'' {\em IEEE Robotics and Automation Letters},
  vol.~5, no.~1, pp.~16--23, 2019.

\bibitem{wnuk2020kinematic}
M.~Wnuk, C.~Hinze, A.~Lechler, and A.~Verl, ``Kinematic multibody model
  generation of deformable linear objects from point clouds,'' in {\em 2020
  IEEE/RSJ International Conference on Intelligent Robots and Systems (IROS)},
  pp.~9545--9552, IEEE, 2020.

\bibitem{song2002potential}
P.~Song and V.~Kumar, ``A potential field based approach to multi-robot
  manipulation,'' in {\em Proceedings 2002 IEEE International Conference on
  Robotics and Automation (Cat. No. 02CH37292)}, vol.~2, pp.~1217--1222, IEEE,
  2002.

\bibitem{bradski2000opencv}
G.~Bradski, ``The opencv library,'' {\em Dr Dobb's J. Software Tools}, vol.~25,
  pp.~120--125, 2000.

\bibitem{long2015fully}
J.~Long, E.~Shelhamer, and T.~Darrell, ``Fully convolutional networks for
  semantic segmentation,'' in {\em Proceedings of the IEEE conference on
  computer vision and pattern recognition}, pp.~3431--3440, 2015.

\bibitem{simonyan2014very}
K.~Simonyan and A.~Zisserman, ``Very deep convolutional networks for
  large-scale image recognition,'' {\em arXiv preprint arXiv:1409.1556}, 2014.

\bibitem{zhang1984fast}
T.~Y. Zhang and C.~Y. Suen, ``A fast parallel algorithm for thinning digital
  patterns,'' {\em Communications of the ACM}, vol.~27, no.~3, pp.~236--239,
  1984.

\bibitem{achlioptas2018learning}
P.~Achlioptas, O.~Diamanti, I.~Mitliagkas, and L.~Guibas, ``Learning
  representations and generative models for 3d point clouds,'' in {\em
  International conference on machine learning}, pp.~40--49, PMLR, 2018.

\end{thebibliography}
\end{document}